\documentclass[10pt,twocolumn,letterpaper]{article}

\usepackage[accsupp]{axessibility}  %

\usepackage[pagenumbers]{cvpr} %

\usepackage{booktabs}
\usepackage{multirow}
\usepackage{adjustbox}
\usepackage{colortbl}
\usepackage{subcaption}
\usepackage{makecell}
\usepackage{tikz}
\usepackage{etoolbox}
\usepackage{threeparttable}
\usepackage{xspace}
\usepackage{bbm}
\usepackage{array}
\usepackage{tabularray}
\usepackage{mfirstuc}

\definecolor{darkblue}{rgb}{0,0,0.5}
\definecolor{darkred}{rgb}{0.8,0,0}
\definecolor{darkgreen}{rgb}{0,0.8,0}

\definecolor{orangered}{HTML}{FF5733}
\definecolor{steelblue}{HTML}{4682B4}
\definecolor{moderategreen}{HTML}{A3CF62}

\definecolor{swgreen}{rgb}{0, 0.6823, 0.7215}
\newcolumntype{a}{>{\columncolor{swgreen! 10}}c}
\newcolumntype{e}{@{\hspace{3pt}}c}
\newcolumntype{f}{@{\hspace{-3pt}}c}

\newcommand{\DL}{\mathcal{D}_\mathcal{L}}
\newcommand{\DU}{\mathcal{D}_\mathcal{U}}
\newcommand{\DUL}{\mathcal{D}_{\mathcal{U}_\mathcal{L}}}
\newcommand{\DUN}{\mathcal{D}_{\mathcal{U}_\mathcal{N}}}
\newcommand{\YL}{\mathcal{Y}_\mathcal{L}}
\newcommand{\YN}{\mathcal{Y}_\mathcal{N}}

\newcommand{\YNP}{\hat{\mathcal{Y}}_\mathcal{N}}

\newcommand{\newparagraph}{\vspace{2pt}\noindent}

\graphicspath{{.}{..}}

\usepackage{dsfont}
\usepackage{pifont}

\newcommand{\methodClip}{CLIP-ZS\xspace}
\newcommand{\methodClipLlm}{CLIP-ZS + LLM vocab\xspace}

\newcommand{\methodClipGT}{CLIP-ZS + GT vocab\xspace}
\newcommand{\methodSSKmeans}{SS-KMeans\xspace}
\newcommand{\methodGcd}{GCD\xspace}
\newcommand{\methodOurs}{\sname (ours)\xspace}

\newcommand{\keyword}[1]{{\it #1}}
\newcommand{\stdv}[1]{{\footnotesize $\pm$#1}}

\usepackage{lipsum}

\usepackage{titletoc}
\usepackage[toc]{appendix}
\usepackage{etoc}
\usepackage{minitoc}
\etocsettocstyle{\section*{Contents}}{}

\newcommand{\context}[1]{\makefirstuc{\textit{#1}}}

\definecolor{cvprblue}{rgb}{0.21,0.49,0.74}
\usepackage[pagebackref,breaklinks,colorlinks,allcolors=cvprblue]{hyperref}

\usepackage[flushmargin,hang,para]{footmisc}

\def\paperID{13712} 
\def\confName{CVPR}
\def\confYear{2025}

\newcommand{\papertitle}{%
Open Ad-hoc Categorization with Contextualized Feature Learning
}

\title{\papertitle}

\newcommand{\lname}{\underbar{O}pen \underbar{A}d-hoc \underbar{C}ategorization with Contextualized Feature Learning\xspace}
\newcommand{\sname}{OAK\xspace}

\author{%
Zilin Wang
\hspace{-2pt}\thanks{
Equal Contribution.  
\url{https://github.com/Wayne2Wang/OAK}
}$\:\:^1$\qquad
Sangwoo Mo\footnotemark[1]$\:\:^1$\qquad
Stella X. Yu$^{1,2}$\qquad
Sima Behpour$^3$\qquad
Liu Ren$^3$
\\
$^1$University of Michigan\qquad
$^2$UC Berkeley\qquad
$^3$Bosch Center for AI
\\
{\small\tt
\{zilinwan,\,swmo,\,stellayu\}@umich.edu\qquad
sima.behpour@gmail.com
\qquad
liu.ren@us.bosch.com
}
\\
}

\begin{document}
\maketitle

\etocdepthtag.toc{mtchapter}
\etocsettagdepth{mtchapter}{subsection}
\etocsettagdepth{mtappendix}{none}
\faketableofcontents

\begin{abstract}
Adaptive categorization of visual scenes is essential for AI agents to handle changing tasks.  Unlike fixed common categories for plants or animals, {\it ad-hoc categories} are created dynamically to serve specific goals. We study {\it open ad-hoc categorization}: Given a few labeled exemplars and abundant unlabeled data, the goal is to discover the underlying context and to expand ad-hoc categories through semantic extension and visual clustering around it. 

Building on the insight that ad-hoc and common categories rely on similar perceptual mechanisms, we propose \sname, a simple model that introduces a small set of learnable context tokens at the input of a frozen CLIP and optimizes with both CLIP’s image-text alignment objective and GCD’s visual clustering objective.

On Stanford and Clevr-4 datasets, \sname achieves state-of-the-art in accuracy and concept discovery across multiple categorizations, including 87.4\% novel accuracy on Stanford Mood, surpassing CLIP and GCD by over 50\%. Moreover, \sname produces interpretable saliency maps, focusing on hands for Action, faces for Mood, and backgrounds for Location, promoting transparency and trust while enabling adaptive and generalizable categorization.
\end{abstract}

\section{Introduction}\label{sec:intro}

\begin{figure}[t]
\definecolor{myred}{HTML}{FF2200}

\definecolor{mygreen}{RGB}{0 200 0}
\definecolor{myblue}{RGB}{0,162,220}
\definecolor{myorange}{RGB}{255,128,0}
\definecolor{myyellow}{RGB}{248,186,0}

\centering
\includegraphics[width=\linewidth]{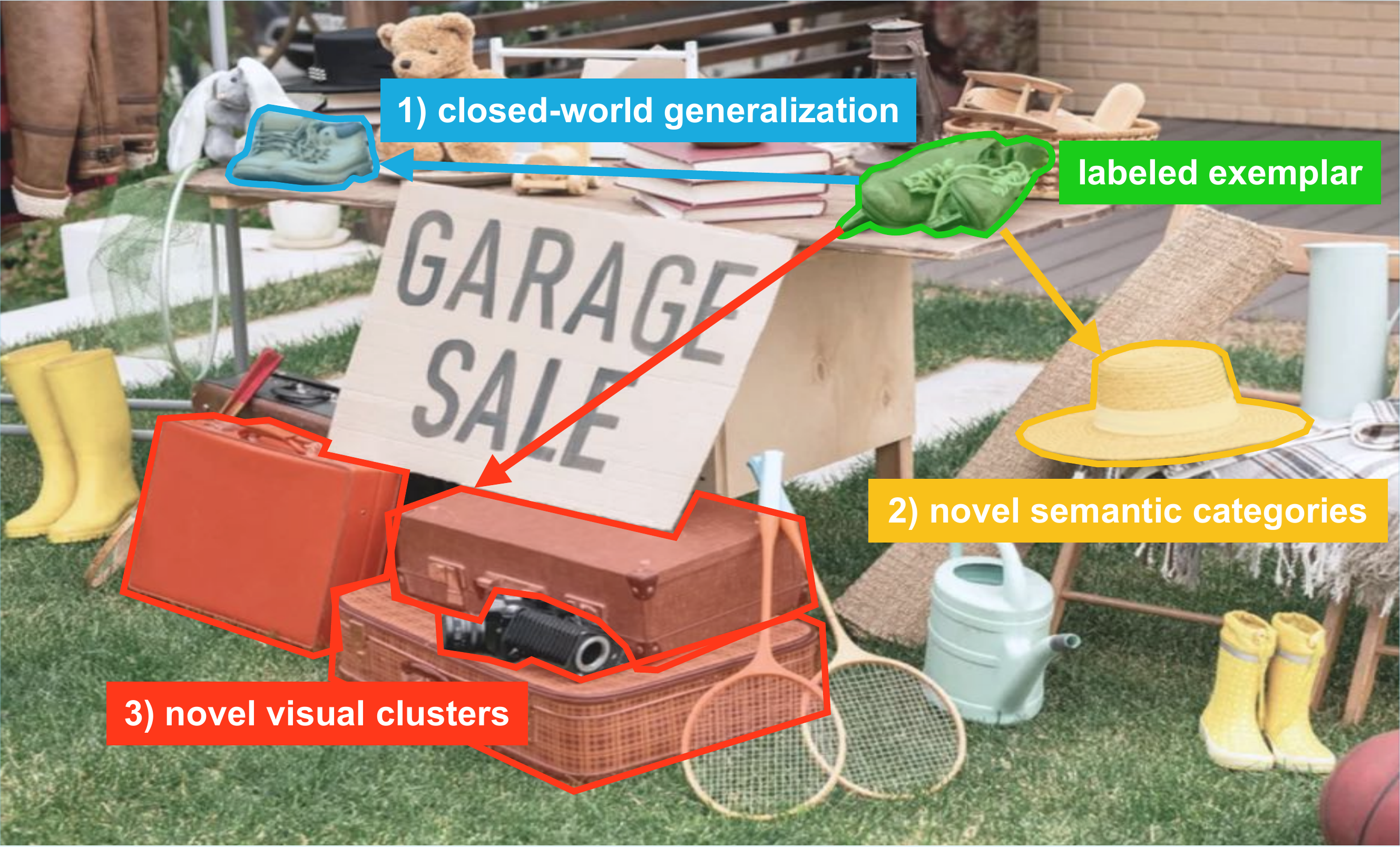}

\caption{%
{\bf We study open ad-hoc categorization} such as \textit{things to sell at a garage sale} to achieve a specific goal ({\it selling unwanted items}).
Given the context {\it garage sale},
{\color{mygreen}labeled exemplars} such as {\color{mygreen}\emph{shoes}}, we need to
recognize all items in the scene that {\it can be sold at the garage sale}, including novel ones.
Supervised models like CLIP focus on
{\color{myblue}1) closed-world generalization}, recognizing other {\color{myblue}\emph{shoes}}.
{\color{myyellow}2) Novel semantic categories} can be discovered by contextual expansion from {\color{mygreen}\emph{shoes}} to {\color{myyellow}\emph{hats}}.
Unsupervised methods like GCD discover
{\color{myred}3) novel visual clusters}, identifying {\color{myred}\emph{suitcases}}.
Our work unifies these scenarios by discovering the latent context and expanding categories both semantically and visually around it.\\[-10pt]
}\label{fig:teaser}
\end{figure}

The concept of {\it ad-hoc categories} in cognitive science differs from {\it common categories} used for plants or animals~\citep{barsalou1983ad}. For example, {\it things to sell at a garage sale} (\cref{fig:teaser}) may lack visual or semantic similarity yet are grouped together to achieve the goal of {\it selling unwanted items}.  Unlike {\it common categories}, {\it ad-hoc categories} are less established in human memory, often lack clear labels, require explicit naming of exemplars, and depend strongly on context. Recognizing them relies on the same perceptual mechanisms as common categories but additionally requires contextualization to adapt to varying goals and situations~\citep{lakoff2008women,rosch1978principles}.

Translating this cognitive notion into machine perception motivates a novel problem setting we call 
\textit{open ad-hoc categorization}, which learns to organize visual data under contexts that are not predefined but must be discovered from a few labeled exemplars and abundant unlabeled data. The context specifies the organizing principle (e.g., {\it action} or {\it location}) and induces distinct category structures (e.g., {\it drinking} vs.\ {\it reading}, {\it urban} vs.\ {\it natural}).  

This task has two objectives: to infer the latent context {\it and} to expand ad-hoc categories accordingly, both semantically by inferring new concept names and visually by identifying coherent clusters. Unlike open-world recognition, where the context is broad and unconstrained, open ad-hoc categorization operates under a specific contextual goal that tightly guides ad-hoc category formation. It leverages a few labeled exemplars to ground this context and reveal task-relevant structure within abundant unlabeled data.

\begin{figure*}[t]

\definecolor{mygreen}{RGB}{0 200 0}
\definecolor{myblue}{RGB}{0,162,220}
\definecolor{myorange}{RGB}{255,128,0}
\definecolor{myyellow}{RGB}{248,186,0}
\definecolor{myred}{HTML}{FF2200}

\centering
\includegraphics[width=\linewidth]{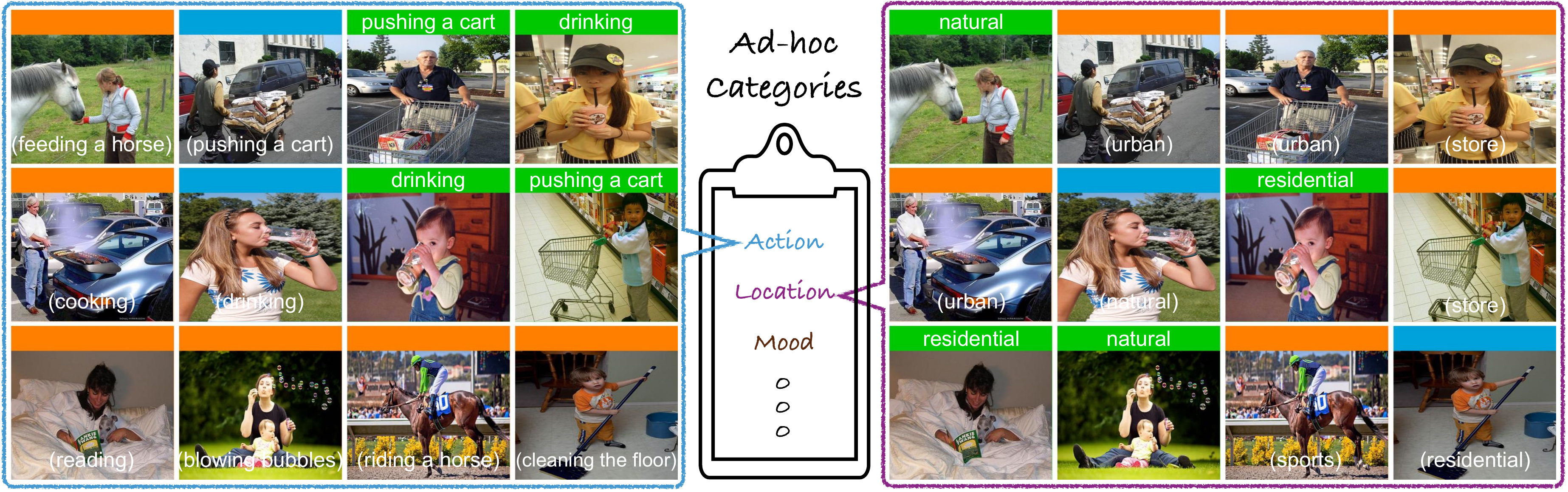}
\caption{%
\textbf{Open ad-hoc categorization} learns diverse categorization rules, dynamically adapting to varying user needs at hand. The same image could be recognized differently depending on the context, such as \emph{drinking} for \context{action} and \emph{residential} for \context{location}. We emphasize the ability to switch between multiple contexts in \sname.
Specifically, given \textbf{1)} a context defined by classes, \textbf{2)} a few labeled images, and \textbf{3)} a set of unlabeled images, \sname holistically reasons over {\color{mygreen}labeled} and unlabeled images, spanning both {\color{myblue}known} and {\color{myorange}novel} classes, to infer novel concepts and propagate labels across the entire dataset.
We show the class names of labeled images in the color box and unlabeled images inside the parentheses, reflecting that the unlabeled class names are not available, only the images.
The \sname setting introduces challenges beyond generalized category discovery (\methodGcd), requiring adaptation to diverse ad-hoc categorization rules based on context.
}\label{fig:problem}

\end{figure*}

\cref{fig:teaser} illustrates three forms of discovery, each reflecting how context constrains category formation within unlabeled data grounded by labeled exemplars. 
\textbf{1)} \textit{Closed-world generalization} applies known concepts to new instances, such as CLIP~\cite{radford2021learning} recognizing unseen {\it shoes}. 
\textbf{2)} \textit{Novel semantic discovery} extends familiar concepts to related ones, such as inferring {\it hats} from {\it shoes}. 
\textbf{3)} \textit{Novel visual discovery} uncovers new clusters that refine the context, such as GCD~\cite{vaze2022generalized} identifying {\it suitcases} as a new category. 
Together, these levels trace a continuum from closed-world generalization to the emergence of new semantic and visual categories shaped by context and revealed through unlabeled data.

We formulate the problem as predicting both known and novel concepts from unlabeled images using only a few labeled examples within a given context. \cref{fig:problem} illustrates how the same set of images can be interpreted differently depending on the context and thus belong to distinct ad-hoc categories, such as \context{action}, \context{location}, or \context{mood}. 
When the context is {\it Action}, images such as {\it a child drinking in a bedroom} are labeled {\it drinking}, and the task is to identify novel {\it Action} classes such as {\it riding} or {\it climbing} and categorize all unlabeled images.  When the context shifts to {\it Location}, the same image is labeled {\it residential area}, and the task becomes inferring novel {\it Location} types such as {\it sports field} or {\it store}.  The few-shot aspect is critical for practicality, enabling rapid adaptation to new contexts and scaling to many more without extensive human annotation.

We build on two insights. 
{\bf 1)} Recognizing ad-hoc categories relies on the same perceptual mechanisms as common categories but requires contextualized interpretation to adapt to varying goals. 
{\bf 2)} Novel concepts can be discovered either semantically, by expanding contextual semantics, or visually, by clustering similar patterns. 
CLIP~\cite{radford2021learning}, trained on internet-scale image–text pairs, captures both fundamental perceptual mechanisms and rich semantic knowledge of common categories, but it assumes a fixed and globally shared semantic space. 
In open ad-hoc categorization, however, the context is not predefined:  The meaning of terms such as {\it Action} or {\it Location} is embodied in a few labeled exemplars and abundant unlabeled data. 
We therefore do not alter CLIP’s mechanisms but instead learn to capture the contextual semantics implicit in the data.  To complement CLIP’s top-down, text-driven interpretation with bottom-up, data-driven discovery, we incorporate GCD~\cite{vaze2022generalized} for visual clustering, enabling joint semantic and visual concept expansion under the discovered context.

We propose \textit{\lname} (\sname for short, with K from Kategorisierung: Categorization in German), a simple model that represents the learned context through a small set of trainable context tokens introduced at the input of a frozen CLIP. These tokens condition CLIP’s image-text representations on the discovered context, grounding its semantic space without modifying its pretrained mechanisms. \sname is trained jointly with CLIP’s image-text alignment objective and GCD’s visual clustering objective, integrating top-down semantic alignment with bottom-up visual discovery. By switching only the context tokens, \sname derives context-specific representations for each ad-hoc categorization, despite its rich global semantics, and provides a single interpretation that does not adapt to any context.

We first evaluate \sname within individual contexts, reporting results separately for known and novel categories. To assess adaptability across contexts, we introduce \textit{Omni accuracy}, a metric that only counts when the model correctly predicts labels from {\it all contexts} for the same image.  This measure reflects realistic scenarios where AI agents must not only learn each context but also switch representations seamlessly across them~\citep{miller2017plans}, such as a household robot alternating between \context{actions} for assistance, \context{locations} for navigation, and \context{moods} for emotional responsiveness.

We benchmark \sname on the Stanford dataset~\citep{kwon2024image} with \context{action}, \context{location}, and \context{mood} contexts, and on the Clevr-4 dataset~\citep{vaze2023no} with \context{texture}, \context{color}, \context{shape}, and \context{count} contexts. Across all settings, \sname outperforms prior methods that rely solely on either semantic or visual cues. For example, \sname achieves 70\% Omni accuracy on the Stanford dataset, compared to 43\% by \methodClipLlm and 52\% by \methodGcd. Beyond accuracy, \sname’s learned features yield interpretable saliency maps that visualize context switching, highlighting {\it hands} for \context{action}, {\it backgrounds} for \context{location}, and {\it faces} for \context{mood}. Finally, \sname can name discovered novel clusters by aligning them with CLIP’s text features, revealing their underlying visual semantics.

To summarize, we make three major contributions.
\begin{itemize}
\item We introduce the \textit{open ad-hoc categorization} task, which unifies context learning with representation switching.
\item We develop \sname, a simple yet effective model that contextualizes CLIP features and incorporates GCD.
\item \sname outperforms strong baselines while producing interpretable saliency maps and naming novel categories.
\end{itemize}

\section{Related Work}
\label{sec:related}

\textbf{Categorization} is a core problem in computer vision, typically assuming common categories like plants and animals~\citep{russakovsky2015imagenet}. Open ad-hoc categorization differs in two important aspects: 1) an open-set setting, where some categories are unknown during training and must be discovered; 2) a context-dependent setting for ad-hoc categories, where shifts in the latent context can drastically alter the categories of interest, necessitating contextualized visual features.

\newparagraph
\textbf{Open-vocabulary classification} models like CLIP~\citep{radford2021learning} have emerged as universal classifiers, enabled by vision-language alignment on internet-scale image-text pairs. By fixing image features and modifying only the text features, CLIP can classify images within any user-defined classes. Recent works attempt to expand and refine the vocabulary by parsing the vocabulary from text databases ~\citep{conti2023vocabulary,han2023s}, prompting large language models (LLMs) for detailed class descriptions~\citep{menon2023visual,pratt2023does,liu2024democratizing,zhao2024ltgc, Park2025FreeGrainedHR}, and leveraging vision-language models (VLMs) to infer candidate vocabulary~\citep{kwon2024image,kim2024discovering,yao2024multi,luo2024llm,liu2024organizing}. Nevertheless, these methods struggle with rare and novel classes and are limited by a single fixed interpretation of an image, making them less effective for the varying contexts highlighted in our \sname task. In contrast, our method adapts image features and discovers novel classes based on the latent context.

\newparagraph
\textbf{Generalized category discovery} (\methodGcd)~\citep{vaze2022generalized} aims to recognize known classes while discovering novel ones in unlabeled images through semi-supervised clustering in a learned feature space. In a subsequent study, \citet{vaze2023no} shows that no single feature space performs well on diverse and changing contexts, underscoring the need for contextualization and motivating our focus on context switching in our \sname task. Another line of research centers on enhancing visual features~\citep{wen2023parametric, rastegar2023learn, rastegar2024selex, fei2022xcon, pu2023dynamic, zhao2023learning, choi2024contrastive, Yang2023GeneralizedCD, Hao2023CiPRAE, Gu2023ClassrelationKD} and improving clustering methods~\citep{wang2024beyond, Otholt2024GuidedCA, Yang2023BootstrapYO}, relying exclusively on visual cues. To take advantage of the rich semantics from text, some works apply \methodGcd to CLIP~\citep{ouldnoughi2023clip,wang2024get,zheng2024textual, su2024multimodal}, but they either depend on external captions or still rely on a single fixed image interpretation. In contrast, \sname exploits semantic guidance from text and learns independent context tokens for each latent context for effective contextual adaptation. As most prior works inherit \methodGcd's core ideas, we base our methods directly on \methodGcd, considering other improvements orthogonal to our contributions.

\nocite{ruff2018deep,tack2020csi,oliver2018realistic,mo2023ropaws,vaze2022open,liu2019large,yang2024generalized}

\newparagraph
\textbf{Multiple clustering.}
Traditional image clustering often reflects the biases inherent in the feature space employed. Deep clustering methods~\citep{caron2018deep, van2020scan} address this by jointly learning the feature space and cluster assignments, but they still only yield a single clustering for the learned feature. Multiple clustering approaches~\citep{qi2009principled, hu2018subspace, yao2023augdmc, metaxas2023divclust, yu2024multiple} instead discover multiple groupings of the same dataset for different contexts, commonly through data augmentations, diversity regularization, or subspace learning. More recently, text-conditioned multiple clustering frameworks~\citep{yao2024multi, yao2025customized, luo2024llm, kwon2024ictc,li2024image, liu2024organizing} enable steering clustering via natural-language criteria using large language models (LLMs) or vision-language models (VLMs). For example, IC$|$TC~\citep{kwon2024ictc} captions each image with a VLM under a user-defined criterion and then refines the captions with an LLM to assign cluster names. OpenSMC~\citep{liu2024organizing} generalizes this paradigm to automatically discover grouping criteria without explicit user input by prompting LLMs and VLMs with carefully designed instructions. However, they remain influenced by the biases of LLMs and VLMs. In contrast, \sname infers the latent clustering contexts from a few user-provided exemplars and class names, and distills them into context tokens that guide clustering and novel class discovery within each context.

\newparagraph
\textbf{Prompt tuning} adapts pre-trained models to specific tasks through a small set of learnable parameters as extra inputs. Originally developed for Natural Language Processing (NLP), early studies~\cite{Li2021PrefixTuningOC, Lester2021ThePO, Liu2021GPTUT, Liu2021PTuningVP} demonstrate that different prompts can effectively contextualize general-purpose language models for specialized tasks such as summarization or question answering. Visual prompt tuning~\citep{jia2022visual, bahng2022exploring} has also been proven effective across various vision tasks. For the \methodGcd task, SPTNet~\citep{wang2024sptnet} introduces learnable spatial prompts arranged in a rectangular layout around each image patch, serving as a form of data augmentation that encourages attention to foreground object regions. PromptCAL~\citep{zhang2023promptcal} inserts learnable prompts as additional inputs to bootstrap visual affinity relationships between classes. Direct supervision is applied to the prompts, and further network layers are also tuned. In \sname, we contextualize CLIP by learning task-specific context tokens independently for each latent context. These tokens capture task-specific contextual information and enable context-aware feature modulation, all without direct supervision. By learning them in a joint text–image space, \sname extends beyond the vision-only counterpart of prompt tuning.

\newparagraph
\textbf{Register tokens}~\citep{darcet2024vision} are also learnable parameters at the input layer but are usually used when training models from scratch, functioning as information hubs that store position-agnostic information. By offloading this information from patch tokens, they allow the patch features to retain more localized information. In contrast, our context tokens encode task-specific context information and guide visual feature extraction with a frozen backbone, ensuring that the resulting image features adapt effectively to the given context.

\section{Open Ad-hoc Categorization}
\label{sec:method}

Open ad-hoc categorization aims to organize visual data under latent, task-specific contexts that are inferred from a few labeled exemplars and abundant unlabeled data. Our \sname method mirrors human cognition: a shared perceptual backbone modulated by context tokens reflects how humans reuse general perception while contextualizing it for ad-hoc goals. \sname can easily switch contexts by only swapping context tokens independently learned for each context.

\subsection{Problem setup}
\label{subsec:method-setup}

Let $\mathcal{X}$ be the image domain and $c$ the underlying latent context with labeled image data $\DL^c$ from known classes $\YL^c$ and unlabeled image data $\DU^c$ containing both known classes $\YL^c$ and novel classes $\YN^c$. The goal is to classify every unlabeled image from $\DU^c$, maximizing accuracy on its known-class subset $\DUL^c$ and novel-class subset $\DUN^c$. 

The latent context $c$ can be inferred through two main principles. \textbf{1) Top-down text guidance}, where the model uses known class names $\YL^c$ and semantic knowledge to infer context. Open-vocabulary classifiers like CLIP~\citep{radford2021learning} can adapt to arbitrary contexts by adjusting their vocabulary. To infer potential novel classes $\YNP^c$, large language models (LLMs) can be prompted with known class names $\YL^c$. CLIP can then predict all classes in this expanded vocabulary $\YL^c \cup \YNP^c$. However, while this text-guided approach accommodates various contexts, it is limited by the pretrained knowledge of CLIP and LLMs. It also does not adapt the image encoder to specific contexts, restricting its representational capacity. \textbf{2) Bottom-up visual clustering}, where the model infers context by grouping visually similar images from the whole dataset $\mathcal{D}^c$. In this approach, GCD~\citep{vaze2022generalized} can be applied to each context to identify categories. However, without semantic cues in the framework, GCD may struggle with complex ad-hoc categories that lack clear visual similarity. It also often requires extensive labeled data, which may be impractical for many ad-hoc categories.

\begin{figure}[t]
\centering
\includegraphics[width=\linewidth]{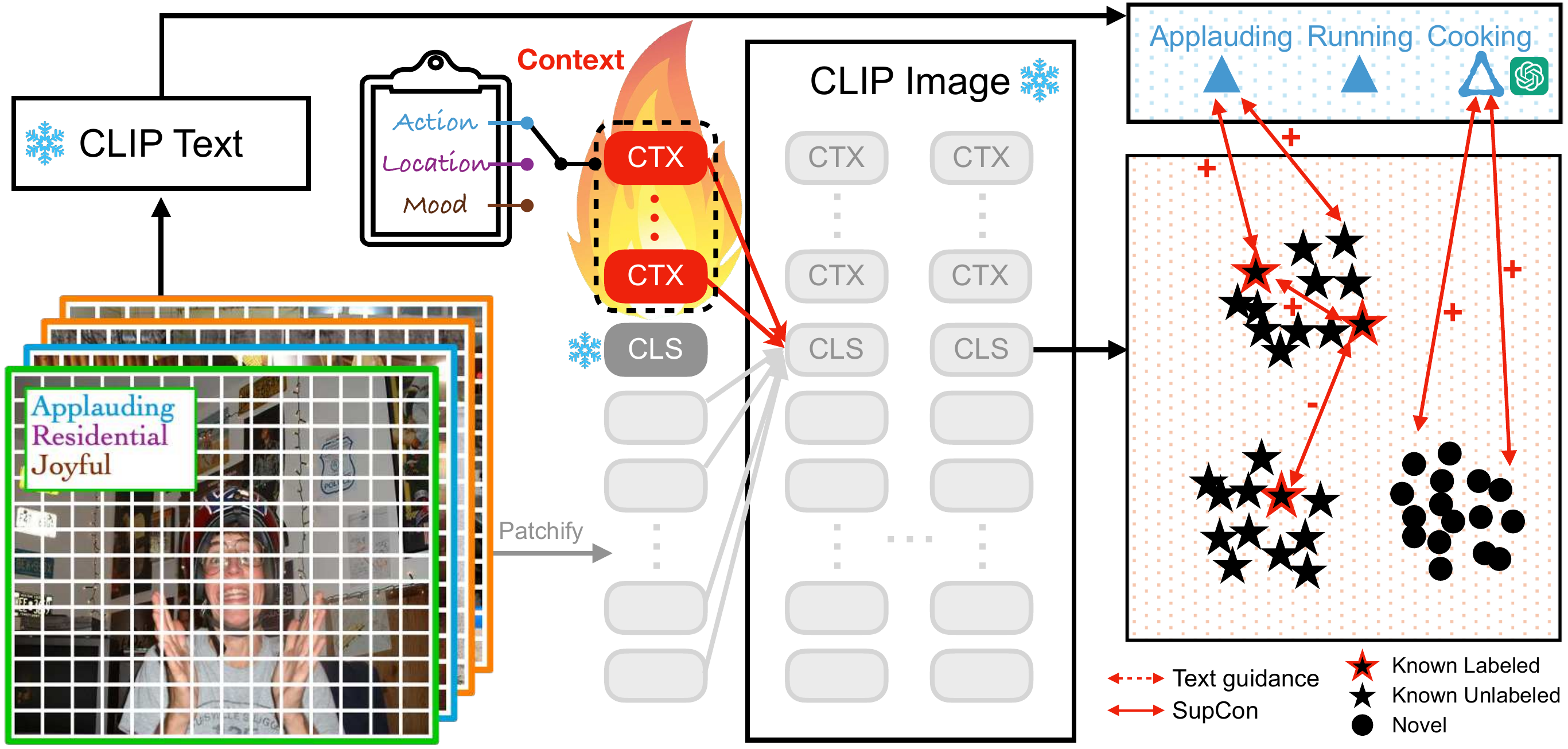}
\vspace{-11pt}
\caption{%
\textbf{\sname learns contextualized features} while preserving the foundations of perception of CLIP
by introducing context tokens that modulate the frozen ViT encoder,
achieving context-aware attention.
This contextualized feature learning follows two key principles: 1) top-down text guidance, which aligns visual clusters with semantic cues and refines clusters accordingly, and 2) bottom-up image clustering, which captures similarity based on visual cues and known class labels. This unified approach effectively combines the individual strengths of CLIP and GCD.
}\label{fig:method}
\end{figure}

\subsection{Our \sname framework}
\label{subsec:method-ours}

We propose \textit{\lname} (\sname), which combines these two principles through three key techniques: \textbf{1)} context-aware visual attention, which adapts pretrained visual embeddings through a small set of context tokens, \textbf{2)} bottom-up visual clustering, which captures visual similarity to infer categorization rules~(GCD), and \textbf{3)} top-down text guidance, which leverages semantic knowledge from class names~(CLIP). This design integrates visual and semantic cues across labeled and unlabeled data.

\newparagraph
\textbf{Background.} CLIP~\citep{radford2021learning} trains an image encoder $f$ and a text encoder $g$ to align embeddings in a shared space on large-scale image–text pairs. The image encoder adopts the ViT~\citep{dosovitskiy2021image} architecture, which divides an image into patches and applies self-attention~\citep{vaswani2017attention} across spatial tokens. The resulting attention map naturally highlights salient regions, guiding the encoder's focus. We employ CLIP as our base visual backbone to exploit its rich semantic knowledge.

\newparagraph
\textbf{Context-aware visual attention.}
Image features should adapt to context, guiding the encoder’s attention toward relevant regions, e.g., focusing on the foreground for \context{actions} and the background for \context{locations}. We introduce context tokens $\mathbf{z}_c$ as additional inputs to the ViT alongside image patch tokens, yielding a context-specific encoder $f_c$:
\begin{equation}
f_c(\mathbf{x}_i) := f([\mathbf{x}_i, \mathbf{z}_c]).
\label{eq:ctx-token}
\end{equation}
These tokens are analogous to register tokens~\citep{darcet2024vision} but tailored and optimized for each context's categorization rule while the backbone remains frozen, as common in visual prompt tuning~\citep{jia2022visual,bahng2022exploring}. This enables adaptation to any ad-hoc categorization by varying only the context tokens.

\newparagraph
\textbf{Bottom-up visual clustering.} 
Clustering visual embeddings $f(\mathbf{x}_i)$ naturally reveals new classes. GCD improves this by training embeddings to group visually and semantically similar images using self-supervised and supervised contrastive losses $\ell_\text{self-con}$ and $\ell_\text{sup-con}$~\citep{wu2018unsupervised,khosla2020supervised} on the unlabeled data $\mathcal{D_U}$ and the labeled data $\mathcal{D_L}$ respectively:
\begin{align}
\ell_\text{GCD}(\mathbf{z})\!=\!
(1\!-\!\lambda)\,\ell_\text{self-con}(\mathbf{z};\mathcal{D_U}) +
\lambda\,\ell_\text{sup-con}(\mathbf{z};\mathcal{D_L}),
\label{eq:loss-gcd}
\end{align}
where $\mathbf{z}$ are trainable parameters and $\lambda$ is a weighting factor. 
While we use GCD for simplicity, advanced variants such as $\mu$GCD~\citep{vaze2023no} are also compatible.

\begin{table*}[t!]
    \caption{%
    \textbf{\sname consistently outperforms baselines, particularly on novel classes and prediction consistency. } We present per-context accuracies and across-context Omni accuracies on \textbf{a)} the Stanford \context{action}, \context{location}, \context{mood} dataset and \textbf{b)} the Clevr-4 \context{texture}, \context{color}, \context{shape}, \context{count} dataset, comparing \sname and open-vocabulary classification (row group 1) and visual clustering (row group 2) baselines, with best results in bold. The advantage is most pronounced in less familiar contexts like Mood. CLIP-ZS + LLM vocab performs poorly on novel classes, revealing the limitation of using class names alone. GCD addresses this by clustering visual features, but \sname goes further by contextualizing them with CLIP’s semantic knowledge, achieving a 50\% gain over both CLIP and GCD in Mood. In the across-context setting, \sname also outperforms all baselines by large margins while demonstrating consistency across contexts. %
     }\label{table:exp}
    \begin{minipage}{\linewidth}
    \centering\small
    \resizebox{1.0\linewidth}{!}{
    \begin{threeparttable}
    \begin{tabular}[b]{l ccc>{\color{lightgray}}a ccca ccca}
    \toprule
        \multicolumn{1}{l}{\textbf{a) Stanford}} &
        \multicolumn{4}{c}{Known} &
        \multicolumn{4}{c}{Novel} &
        \multicolumn{4}{c}{Overall} \\
        \cmidrule(lr){2-5} \cmidrule(lr){6-9} \cmidrule(lr){10-13}
        Method &
        Action & Location & Mood & Omni\tnote{a} &
        Action & Location & Mood & Omni &
        Action & Location & Mood & Omni \\
    \midrule
    \methodClip &
    \textbf{96.2} & 60.4 & \textbf{87.4} & 75.0 &
    - & - & - & - &
    - & - & - & - \\
    + LLM vocab &
    93.5 & 58.3 & 75.9 & 75.0 &
    38.6 & 34.2 & 35.4 & \phantom{0}0.0 &
    65.2 & 47.5 & 55.0 & 43.0 \\
    + GT vocab &
    93.6 & 59.6 & 78.8 & 75.0 &
    80.3 & 59.7 & 65.8 & 29.4 &
    86.7 & 59.7 & 72.1 & 38.3 \\
    \midrule
    \methodSSKmeans &
    67.0 & 56.3 & 43.2 & \phantom{0}0.0 &
    53.9 & 71.1 & 78.2 & 35.3 &
    60.2 & 62.9 & 61.3 & 47.7 \\
    \methodGcd &
    89.4 & 75.4 & 64.3 & 75.0 &
    67.8 & 80.8 & 40.6 & \phantom{0}0.0 &
    78.3 & 77.8 & 52.1 & 52.3 \\
    \midrule
    \sname (ours) &
    88.9 & \textbf{83.9} & 68.8 & \phantom{0}0.0 &
    \textbf{85.1} & \textbf{88.4} & \textbf{87.4} & \textbf{47.1} &
    \textbf{86.9} & \textbf{85.9} & \textbf{78.4} & \textbf{70.3} \\
    \bottomrule
    \end{tabular}
    \begin{tablenotes}
    \footnotesize
        \item[a] \color{lightgray} With only 8 images overlapping across all contexts, text is toned down to light gray due to unreliable results.
    \end{tablenotes}
    \end{threeparttable}}
    \end{minipage}

    \vspace{5pt}

    \begin{minipage}{\linewidth}
    \centering\large
    \resizebox{1.0\linewidth}{!}{
    \begin{tabular}[b]{l cccca cccca cccca}
    \toprule
        \multicolumn{1}{l}{\textbf{b) Clevr-4}} &
        \multicolumn{5}{c}{Known} &
        \multicolumn{5}{c}{Novel} &
        \multicolumn{5}{c}{Overall} \\
        \cmidrule(lr){2-6} \cmidrule(lr){7-11} \cmidrule(lr){12-16}
        Method &
        Texture & Color & Shape & Count & Omni &
        Texture & Color & Shape & Count & Omni &
        Texture & Color & Shape & Count & Omni \\
    \midrule
    \methodClip &
    40.8 & 94.4 & 79.7 & \textbf{46.9} & 14.6 &
    - & - & - & - & - &
    - & - & - & - & - \\
    + LLM vocab &
    29.4 & 81.9 & 56.5 & 21.7 & \phantom{0}3.3 &
    20.7 & 44.1 & 49.5 & 24.4 & \phantom{0}0.6 &
    25.0 & 62.8 & 53.3 & 23.1 & \phantom{0}1.7 \\    
    + GT vocab &
    13.3 & 79.8 & 60.1 & 21.7 & \phantom{0}3.1 &
    26.5 & 62.7 & 64.8 & 24.4 & \phantom{0}2.4 &
    20.0 & 71.2 & 62.3 & 23.1 & \phantom{0}2.0 \\
    \midrule
    \methodSSKmeans &
    12.9 & 10.4 & 73.2 & 24.2 & \phantom{0}0.5 &
    13.7 & 13.0 & 82.8 & 15.5 & \phantom{0}0.2 &
    13.4 & 11.7 & 77.6 & 19.8 & \phantom{0}0.1 \\
    \methodGcd &
    73.4 & 98.3 & 99.0 & 41.9 & 35.5 &
    43.6 & 94.9 & 99.2 & 42.3 & 15.7 &
    58.2 & 96.6 & 99.1 & 42.1 & 22.6 \\
    \midrule
    \sname (ours) &
    \textbf{82.3} & \textbf{100.0} & \textbf{99.9} & 45.0 & \textbf{40.5} &
    \textbf{47.8} & \textbf{100.0} & \textbf{99.8} & \textbf{43.7} & \textbf{16.5} &
    \textbf{64.6} & \textbf{100.0} & \textbf{99.8} & \textbf{44.4} & \textbf{28.5}  \\
    \bottomrule
    \end{tabular}
    }
    \end{minipage}

\end{table*}

\begin{figure*}[t]
\definecolor{myred}{RGB}{255,87,51}
\definecolor{myblue}{RGB}{83,127,176}
\definecolor{mygreen}{RGB}{72,205,112}
\small\centering
\newcommand{\squareimage}[1]{\includegraphics[width=0.136\linewidth, height=0.136\linewidth]{#1}}
\newcommand{\colspace}{\hspace{4pt}}
\newcommand{\colspaceclose}{\hspace{2pt}}
\newcommand{\tablerow}[1]{
\squareimage{figs/saliency/stanford_#1.png} &
\squareimage{figs/saliency/stanford_#1_clip.png} &
\squareimage{figs/saliency/stanford_#1_gcd_action.png} &
\squareimage{figs/saliency/stanford_#1_cdc_action.png} &
\squareimage{figs/saliency/stanford_#1_gcd_location.png} & 
\squareimage{figs/saliency/stanford_#1_cdc_location.png} &
\squareimage{figs/saliency/stanford_#1_gcd_mood.png} & 
\squareimage{figs/saliency/stanford_#1_cdc_mood.png}
}
\resizebox{1.0\linewidth}{!}{
\begin{tabular}
{@{}
c@{\colspace}c@{\colspace}
c@{\colspaceclose}c@{\colspace}
c@{\colspaceclose}c@{\colspace}
c@{\colspaceclose}
c@{}}

\multicolumn{2}{l}{\textbf{a) Stanford}} &
\multicolumn{2}{c}{Action} &
\multicolumn{2}{c}{Location} &
\multicolumn{2}{c}{Mood} \\
\cmidrule(lr){3-4}\cmidrule(lr){5-6}\cmidrule(lr){7-8}
Image & 
CLIP &
GCD & \sname (ours) &
GCD & \sname (ours) &
GCD & \sname (ours) \\
\tablerow{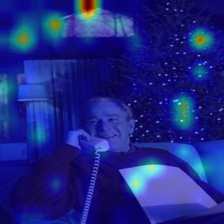}\\[-1pt]
& & %
\textcolor{darkgreen}{phoning} & \textcolor{darkgreen}{phoning} & 
\textcolor{darkred}{market} & \textcolor{darkgreen}{residential} & 
\textcolor{darkred}{focused} & \textcolor{darkgreen}{relaxed} \\[1pt]
\tablerow{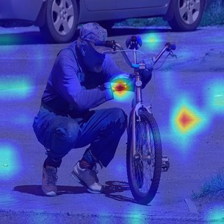}\\[-1pt]
& & %
\textcolor{darkred}{riding a bike} & \textcolor{darkgreen}{fixing a bike} & 
\textcolor{darkred}{transportation} & \textcolor{darkgreen}{city street} & 
\textcolor{darkred}{joyful} & \textcolor{darkgreen}{focused} \\
\end{tabular}

}

\vspace{1pt}

\renewcommand{\squareimage}[1]{\includegraphics[width=0.105\linewidth, height=0.105\linewidth]{#1}}
\renewcommand{\tablerow}[1]{
\squareimage{figs/saliency/clevr4_#1.png} & 
\squareimage{figs/saliency/clevr4_#1_clip.png} & 
\squareimage{figs/saliency/clevr4_#1_gcd_texture.png} &
\squareimage{figs/saliency/clevr4_#1_cdc_texture.png} &
\squareimage{figs/saliency/clevr4_#1_gcd_color.png} &
\squareimage{figs/saliency/clevr4_#1_cdc_color.png} &
\squareimage{figs/saliency/clevr4_#1_gcd_shape.png} &
\squareimage{figs/saliency/clevr4_#1_cdc_shape.png} &
\squareimage{figs/saliency/clevr4_#1_gcd_count.png} &
\squareimage{figs/saliency/clevr4_#1_cdc_count.png} 
}

\resizebox{1.0\linewidth}{!}{
\begin{tabular}
{@{}
c@{\colspace}c@{\colspace}
c@{\colspaceclose}c@{\colspace}
c@{\colspaceclose}c@{\colspace}
c@{\colspaceclose}c@{\colspace}
c@{\colspaceclose}
c@{}}

\multicolumn{2}{l}{\textbf{b) Clevr-4}} &
\multicolumn{2}{c}{Texture} &
\multicolumn{2}{c}{Color} &
\multicolumn{2}{c}{Shape} &
\multicolumn{2}{c}{Count} \\
\cmidrule(lr){3-4}\cmidrule(lr){5-6}\cmidrule(lr){7-8}\cmidrule(lr){9-10}
Image & CLIP &
GCD & \sname (ours) &
GCD & \sname (ours) &
GCD & \sname (ours) &
GCD & \sname (ours) \\
\tablerow{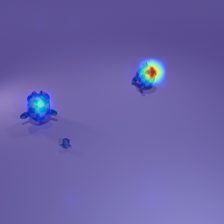}\\[-1pt]
& & %
\textcolor{darkred}{circle} & \textcolor{darkgreen}{checkered} & 
\textcolor{darkgreen}{orange} & \textcolor{darkgreen}{orange} & 
\textcolor{darkgreen}{monkey} & \textcolor{darkgreen}{monkey} &
\textcolor{darkred}{2} & \textcolor{darkgreen}{3} \\[1pt]
\tablerow{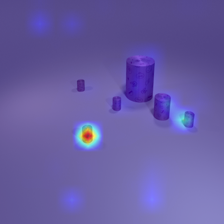}\\[-1pt]
& & %
\textcolor{darkred}{chessboard} & \textcolor{darkgreen}{emojis} & 
\textcolor{darkgreen}{purple} & \textcolor{darkgreen}{purple} & 
\textcolor{darkgreen}{cylinder} & \textcolor{darkgreen}{cylinder} &
\textcolor{darkred}{5} & \textcolor{darkgreen}{6} \\
\end{tabular}

}

\caption{
\textbf{\sname attends to context-relevant regions of images, while CLIP and \methodGcd often focus on arbitrary or less informative areas.} We present saliency maps on \textbf{a)} the Stanford \context{action}, \context{location}, \context{mood} dataset and \textbf{b)} the Clevr-4 \context{texture}, \context{color}, \context{shape}, \context{count} dataset. We visualize saliency maps for CLIP, GCD, and \sname using the method of \citet{chefer2021transformer}, guided by the predicted class, except for CLIP, where an empty string is used. Predictions are color-coded as \textcolor{darkgreen}{correct} and \textcolor{darkred}{incorrect}. On the Stanford dataset, \sname highlights human behaviors such as \textit{hand movements} for \context{action}, captures the \textit{entire scene} for \context{location}, and emphasizes the \textit{human face} for \context{mood}, aligning well with human intuition. While GCD produces reasonable maps for some \context{action} examples, like \keyword{phoning}, it fails in cases like \keyword{fixing a bike}, mistakenly attending to the bike rather than the human action, confusing it with \keyword{riding a bike}. CLIP, on the other hand, consistently focuses on salient objects like \keyword{humans} but does not adapt its attention to different contexts.
}\label{fig:vis}
\vspace{-8pt}
\end{figure*}

\newparagraph
\textbf{Top-down text guidance.}
The GCD objective clusters images by class but ignores semantic relations, treating classes as independent entities. To address this, we align clusters with pretrained text embeddings by freezing the text encoder $g$ and fine-tuning only the image encoder $f$ using context tokens $\mathbf{z}_c$. We apply a $|\YL \cup \YNP|$-way classification loss between image and text embeddings, where $\YNP$ denotes potential novel classes generated by prompting LLMs, though other methods such as textual inversion~\citep{wen2024hard} may also be used. For labeled data $\DL$, we use ground-truth labels; for unlabeled data $\DU$, pseudo-labels are derived from semi-supervised K-means (SS-KMeans) with Hungarian matching~\citep{kuhn1955hungarian} between cluster and text embeddings at each training epoch.
Formally, our text guidance loss is:
\begin{equation}
\begin{aligned}
\ell_\text{text}(\mathbf{z}_c)
&= \frac{1}{|\DL|} \sum_{(\mathbf{x}_i, y_i) \in \DL}
\mathrm{CE}\left( p(y_i | \mathbf{x}_i; \mathbf{z}_c), y_i \right)\\
&\hspace{9pt}+ \frac{1}{|\DU|} \sum_{\mathbf{x}_i \in \DU}
\mathrm{CE}\left( p(\hat{y}_i | \mathbf{x}_i; \mathbf{z}_c), \hat{y}_i \right)
\label{eq:loss-text}
\end{aligned}
\end{equation}
where $p(y_i | \mathbf{x}_i; \mathbf{z}_i)$ is the $i$-th value of the softmax probability over the cosine similarities between the image embedding $f([\mathbf{x}_i, \mathbf{z}_c])$ and text embeddings $\{g(y_i) \,|\, y_i \in \mathcal{Y_L}\}$, $\hat{y}_i$ is the pseudo-label for an unlabeled image $\mathbf{x}_i$, and $\mathrm{CE}$ denotes cross-entropy. This semi-supervised approach is similar to GPC~\citep{zhao2023learning}, but we incorporate textual information to obtain pseudo-labels from visual clusters. 

\newparagraph
{\bf Full training objective.} Our total loss for \sname is:
\begin{align}
\ell_\text{\sname}(\mathbf{z}_c) = \ell_\text{GCD}(\mathbf{z}_c) + \lambda_\text{text} \cdot \ell_\text{text}(\mathbf{z}_c)
\label{eq:loss-all}
\end{align}
where $\lambda_\text{text}$ is a hyperparameter for the relative strength of text guidance. In our experiments, we observe that it is most useful when the CLIP image encoder is less familiar with certain concepts, as in \context{location} or \context{mood}, where the text encoder supplies complementary cues.

\newparagraph
\textbf{Inference.} 
After training, we obtain the visual clusters and their pseudo-labels using the same approach as in our semi-supervised learning, providing predictions for both known and novel classes. Unlike GCD, which discovers clusters in the image encoder alone, text guidance allows \sname to assign text labels to each cluster, offering a clearer interpretation of open categories.

\section{Experiments}
\label{sec:exp}

Our \sname method learns from both top-down semantic guidance and bottom-up visual clustering by adapting context-aware visual attention through a small set of context tokens specific to each context. We evaluate its effectiveness on our proposed open ad-hoc categorization tasks, focusing on both per-context accuracy and Omni accuracy across all contexts for both known and novel classes. We compare \sname against two categories of baselines: 1) semantic-only methods, like CLIP~\citep{radford2021learning} with extended vocabularies; and 2) visual-only methods, like GCD~\citep{vaze2022generalized}, highlighting the benefits of joint semantic-visual training and the use of context tokens. We also analyze saliency maps to further illustrate the contextualization process. %

\subsection{Experimental setup}
\label{sec:exp-setup}

{\bf Datasets.}
To benchmark the effectiveness of \sname, we employ two datasets that each come with different contexts. These contexts are latent to the model and must be inferred. The context names are never provided to the model, and we include the context names only for the convenience of reference. 
\textbf{1) The Stanford dataset} ~\citep{kwon2024image} is originally collected for 40 \context{action} classes~\citep{yao2011human} (9.2K images) and later annotated with 10 \context{location} (920 images) and 4 \context{mood} classes (968 images). For each context, we randomly select half of the classes as known and the rest as novel, sampling 16 images per class and treating them as labeled. These sets overlap but are not identical. We use the 127 overlapping images across all contexts to benchmark the Omni accuracy. For the known classes and novel classes, our evaluation sets only have 8 and 17 overlapping images across all contexts in our split.
\textbf{2) The Clevr-4 dataset}~\citep{vaze2023no} is a synthetic dataset rendered in the CLEVR~\citep{johnson2017clevr} environment, featuring various objects in a controlled setting. It contains 8.3K images and 10 classes per context: \context{texture}, \context{color}, \context{shape}, and \context{count}. Similar to the Stanford dataset, half the classes in each context are randomly selected as known, and the rest as novel, with 16 images per class sampled for the labeled set. All images are annotated with all contexts, using the full dataset of 8.3K images to evaluate the overall, known, and novel Omni accuracy. We provide the specific statistics in \cref{supp:dataset-details}.

\newparagraph
\textbf{Baselines.}
We consider two sets of baselines: 1) semantic-only methods for top-down text guidance, such as CLIP~\citep{radford2021learning} with an extended vocabulary, and 2) visual-only methods for bottom-up image clustering, such as GCD~\citep{vaze2022generalized}. For the semantic-only group, we consider the following baselines:

\begin{itemize}
\item \textbf{\methodClip}:
CLIP zero-shot (ZS) classifier relies solely on known class names, serving as a closed-world baseline unable to discover novel categories.

\item \textbf{\methodClipLlm}:
Extend CLIP-ZS to predict novel classes by generating potential class names using large language models (LLMs).

\item \textbf{\methodClipGT}:
Apply CLIP-ZS using ground-truth (GT) class names for novel categories, setting the upper bound for zero-shot methods.
\end{itemize}

\newparagraph
For the visual-only group, we consider the following:

\begin{itemize}
\item \textbf{\methodSSKmeans}:
Extract visual embeddings from CLIP without fine-tuning, then use semi-supervised K-means (SS-KMeans) clustering to discover novel classes.

\item \textbf{\methodGcd}:
Fine-tune CLIP with the GCD loss and use SS-KMeans clustering to discover novel classes.
\end{itemize}

We reimplement all baselines and our method within a consistent experimental setup, using a ViT-B/16 image encoder with the CLIP weights released by OpenAI. Following the GCD training recipe, we adjust only the learning rates and epochs for each dataset. For a fair comparison, the same training procedure is applied across all methods.

\newparagraph
\textbf{Additional baselines.}
We primarily focus on a few-shot setup, as collecting extensive labels for diverse ad-hoc contexts is impractical. Consequently, our reimplemented baselines may underperform compared to values reported in prior works that use the entire labeled sets. For further validation, we also report comparisons with state-of-the-art methods in the full-shot scenario (\cref{table:exp-sota-gcd}), which includes GCD~\citep{vaze2022generalized}, SimGCD~\citep{wen2023parametric}, and $\mu$GCD~\citep{vaze2023no}.

\newparagraph
\textbf{Evaluation metrics.}
Following the standard for generalized category discovery~\citep{vaze2022generalized}, we report accuracy for known ($\mathcal{D_{U_L}}$), novel ($\mathcal{D_{U_N}}$), and overall ($\mathcal{D_U}$) classes. Novel accuracy varies by method: CLIP-ZS uses an extended vocabulary, while GCD applies Hungarian matching. Each method is evaluated independently within each context using its respective labeled set. We also report Omni accuracy, which measures correctness across all contexts. Formally, for images $\{\mathbf{x}_i\}_{i=1}^N$ and contexts $c \in \mathcal{C}$, with $\hat{y}^c$ and $y^c$ as predicted and true labels, the Omni accuracy is:
\begin{align}
\frac{1}{N} \, \sum_{i=1}^N \, \mathbbm{1}\left( \, \bigwedge_{c \in \mathcal{C}} \, \hat{y}^c_i = y^c_i \right).
\end{align}

\newparagraph
\textbf{Assumptions beyond GCD.}
We make no additional assumptions about the data and label distribution beyond GCD. Both methods assume a known number of novel classes, though it can also be estimated by a method outlined in the GCD paper. The key difference lies in our pool of class names, which can be generated by an LLM.

\subsection{Results and analysis}
\label{sec:exp-main-results}
\textbf{Main benchmark results. } %
In \cref{table:exp}, we present the per-context and the Omni accuracies on the Stanford and Clevr-4 datasets. As shown, \sname outperforms all baselines in both novel and overall accuracy, demonstrating the necessity of visual feature contextualization and the effectiveness of our context tokens. Combining the strengths of both CLIP and \methodGcd, \sname also surpasses their performances of they are alone. The closed-world CLIP zero-shot baseline has an unfair advantage in that it only scores against the known classes, which is much smaller than the true label space. This naturally boosts its performance on the known classes and prevents it from discovering novel ones. 
We also find that CLIP is less effective on Clevr-4 than Stanford due to its limited exposure to synthetic images and niche contexts.
Nonetheless, text guidance still benefits \sname due to the rich text understanding from CLIP training, enabling \sname to consistently improve upon \methodGcd.

\newparagraph
\textbf{Saliency maps.}
The label of an image depends on the categorization context, so the model’s focus must adapt accordingly. To validate this, we compare saliency maps from CLIP, GCD, and \sname using the method of \citet{chefer2021transformer}. \cref{fig:vis} a) shows saliency maps on the Stanford dataset, suggesting that \sname aligns closely with human intuition. It focuses on hands for \context{action}, the entire scene for \context{location}, and faces for \context{mood}. All done without direct supervision. In contrast, GCD often attends to irrelevant regions, particularly in \context{location} and \context{mood}, where the domain gap from CLIP is larger, leading to errors. CLIP always focuses on the statistically significant parts regardless of the context.

We further study saliency maps on the Clevr-4 dataset in \cref{fig:vis} b) following the same setup. As shown, \sname adapts its attention based on context complexity, using fewer queries for simple concepts and more for complex ones. For example, it attends to small regions for \context{texture} and \context{color}, attends to multiple 2D views for \context{shape}, and focuses on all objects for \context{count}. In contrast, GCD fails to adjust its attention appropriately, missing key regions such as objects in the Count context, while CLIP attends broadly to all objects and background, regardless of the context.

\begin{table}[t]
    \small
    \caption{%
    \textbf{\sname maintains state-of-the-art performance in the full-shot setting on Clevr-4.} We report novel-class accuracies across Clevr-4 contexts, with baseline numbers taken from \citet{vaze2023no}. \sname excels particularly in the challenging \context{texture} context with an 11\% gain over $\mu$GCD. While \citet{vaze2023no} finds that larger pretrained models are not consistently effective for all contexts and trains ResNet-18~\citep{he2016deep} from scratch, \sname's pretrained ViT~\citep{dosovitskiy2021image} modulated by context tokens performs robustly in all.
    }\label{table:exp-sota-gcd}
    \resizebox{1.0\linewidth}{!}{
    \setlength{\tabcolsep}{6pt}
    \begin{tabular}[b]{@{}l cccc |c @{}}
    \toprule
        Method &
        Texture & Color & Shape & Count & Average \\
    \midrule
    GCD~\citep{vaze2022generalized} &
    45.3 & 90.5 & 88.5 & 60.1 
    & 71.1\\
    SimGCD~\citep{wen2023parametric} &
    40.2 & 97.2 & 95.1 & 53.9
    & 71.6\\
    $\mu$GCD~\citep{vaze2023no} &
    55.5 & 92.1 & \textbf{99.2} & 65.2
    & 78.0\\
    \midrule
    \methodOurs &
    \textbf{66.5} & \textbf{99.9} & 99.0 & \textbf{67.6}
    & \textbf{83.3} \\
    \bottomrule
    \end{tabular}
    }
\vspace{-10pt}
\end{table}

\newparagraph
\textbf{Full-shot results.}
\sname is proven to be effective not only in few-shot ad-hoc categorization but also in full-shot GCD setups. We evaluate on the full Clevr-4 using 2K labels per context. \cref{table:exp-sota-gcd} shows that \sname outperforms the state-of-the-art $\mu$GCD~\cite{vaze2023no} in novel accuracy by a great margin, achieving near-perfect results for \context{color} and \context{shape} and an 11\% gain on \context{texture}. Although $\mu$GCD reports that no single representation generalizes across contexts and that pretrained models fail in abstract domains like Clevr-4, \sname succeeds through unified feature contextualization.

\begin{figure}[t]
\centering
\includegraphics[width=\linewidth]{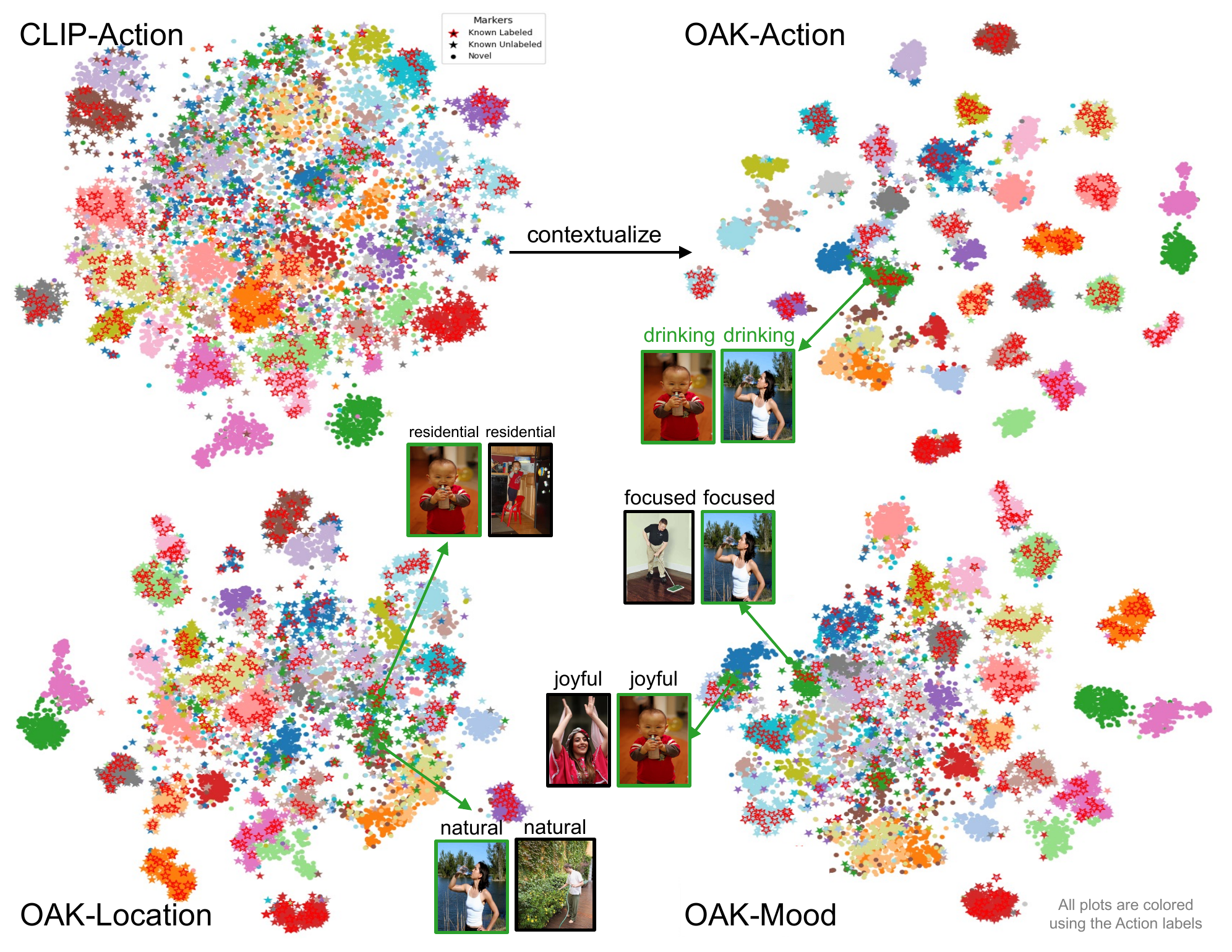}
\caption{%
\textbf{\sname can effectively contextualize and switch between diverse contexts.} We show t-SNE plots of CLIP and \sname with points colored by the ground-truth class in \context{action}: known images as stars, labeled images as red-bordered stars, and novel images as dots. While CLIP shows poor clustering under \context{action}, \sname's contextualized features form well-separated clusters. Additionally, images grouped closely in \sname under \context{action} become far apart in other contexts, underscoring the context-dependent interpretation and the context-switching ability of \sname.
}\label{fig:analysis-tsne}
\vspace{-5pt}
\end{figure}

\newparagraph
\textbf{t-SNE visualizations.}
We use t-SNE~\citep{van2008visualizing} to visualize embeddings from CLIP and \sname across different contexts in the Stanford dataset in \cref{fig:analysis-tsne}. CLIP often attends to the most statistically salient image regions, typically the human, while the \context{action} context requires recognizing both the human hand and the manipulated object. As a result, CLIP’s feature space forms diffuse, poorly separated clusters. In contrast, \sname, explicitly contextualized for \context{action}, produces clear and well-separated clusters. Furthermore, \sname adapts its representation to the context: images that are close together in \context{action} may lie far apart in \context{location} or \context{mood}, illustrating \sname's ability to switch contexts. Detailed per-context plots are available in \cref{subsec:supp-tsne}.

\begin{figure}[t]
\small\centering
\newcommand{\squareimage}[1]{\includegraphics[width=0.3\linewidth, height=0.26\linewidth]{#1}}
\newcommand{\colspace}{\hspace{3pt}}

\begin{tabular}
{@{}c@{\colspace}c@{\colspace}c@{}}
\multicolumn{3}{c}{\keyword{jumping} named as \keyword{dancing}} \\
\squareimage{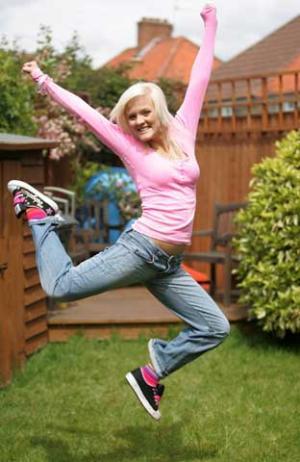} &
\squareimage{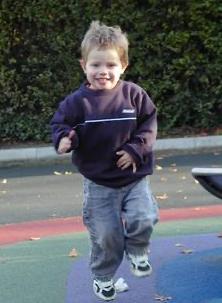} &
\squareimage{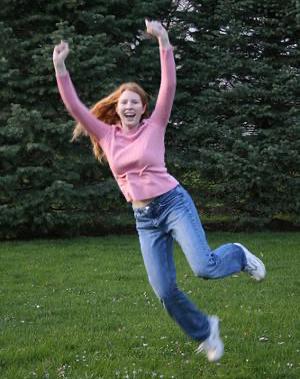}\\
\multicolumn{3}{c}{\keyword{pouring liquid} named as \keyword{carrying a box}} \\
\squareimage{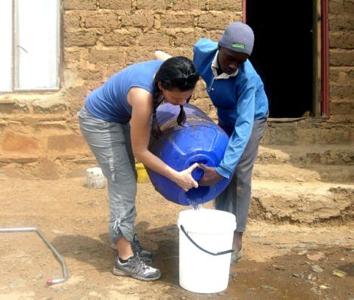} &
\squareimage{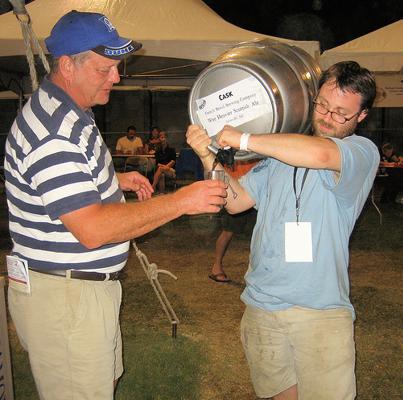} &
\squareimage{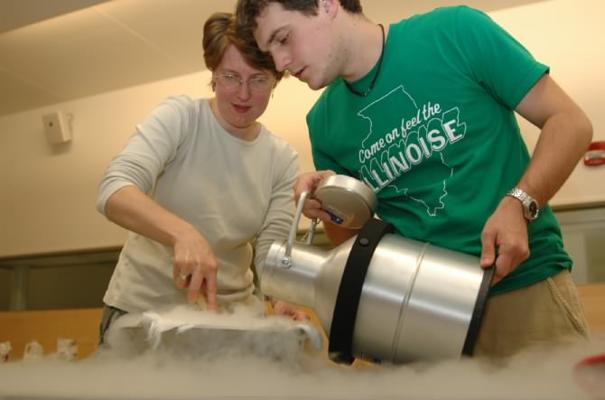} \\
\end{tabular}
\caption{
\textbf{\sname can reasonably name the discovered concepts.} We show visual examples with ground-truth and \sname-predicted class names. \sname handles ambiguous examples well, for instance, predicting \keyword{jumping} for images annotated as \keyword{dancing}, both of which are plausible interpretations under the \context{action} context.
}\label{fig:name-example}
\vspace{-10pt}
\end{figure}

\newparagraph
\textbf{Naming clusters.}
\sname can name visual clusters by matching image and text embeddings. In \cref{fig:name-example}, we visualize interesting predicted class names that differ from annotated labels but are still reasonable, such as images in the \keyword{dancing} cluster appearing like \keyword{jumping} despite their true label. An exhaustive list of true and predicted class names for novel clusters is available in \cref{supp:clustername}. \sname can accurately name novel visual concepts, such as \keyword{preparing a meal} for \keyword{cooking} images. It performs well in contexts familiar to CLIP, like \context{action}, and also in the challenging context of \context{count}, a concept CLIP finds particularly difficult to understand~\citep{paiss2023teaching}.

\section{Summary}

We study open ad-hoc categorization, which predicts both known and novel classes in unlabeled data under varying contexts. We propose \sname, a context-token-based extension of CLIP that integrates visual and semantic cues. This work advances AI agents toward seamlessly handling diverse real-world tasks. Extending ad-hoc categorization beyond static images is an exciting direction for future work.

\newparagraph{\bf Acknowledgments.}
This project was supported, in part, by Bosch gift funds to S. Yu at UC Berkeley and UMich.

\clearpage
{
    \small
    \bibliographystyle{ieeenat_fullname}
    \bibliography{main}
}

\clearpage
\appendix

\onecolumn

\begin{center}{\bf \Large \papertitle}\end{center}
\begin{center}{\Large Supplementary Material}\end{center}

\hypersetup{linkcolor=black}
\etocdepthtag.toc{mtappendix}
\etocsettagdepth{mtchapter}{none}
\etocsettagdepth{mtappendix}{subsection}
\tableofcontents
\hypersetup{linkcolor=cvprblue}

\clearpage
\section{Experimental details}
\label{supp:detail}

\subsection{Dataset details}
\label{supp:dataset-details}

The Stanford \context{action} dataset~\cite{yao2011human} is available from its official website, while the Stanford \context{location} and \context{mood} dataset~\cite{kwon2024image} can be downloaded from its official GitHub page. We generate a text file containing filenames and ground-truth labels for each dataset. In the smaller Stanford \context{location} and \context{mood} dataset, we retain all filenames present in Stanford \context{action} and use a special symbol to indicate missing images. All available images from these datasets are used. For Clevr-4, the datasets \cite{vaze2023no} constructed using the CLEVR environment~\citep{johnson2017clevr} are available on the authors' GitHub page, and we use them without additional preprocessing, utilizing only the training split for our method. We provide the dataset statistics in \cref{table:data}, complete class names in \cref{tab:class-names-supp}, and evaluation set sizes (overlap across all contexts) in \cref{tab:eval-size-supp}. 

\begin{table*}[h]
    \caption{%
    \textbf{Dataset statistics} used in our experiments. We randomly split the classes, assigning half as known ($\YL$) and the other half as novel ($\YN$), sampling 16 images per class for the labeled set ($\DL$) and using the remaining images for the unlabeled set ($\DU$) in each context. This setup reflects the practical scenario of ad-hoc categorization, where obtaining extensive labels for diverse contexts is challenging. Please note that our results are not directly comparable to prior work, which often uses thousands of labeled samples.
    }\label{table:data}
    \centering\small
    \begin{tabular}[b]{l ccc}
    \toprule
    \textbf{a) Stanford} & Action & Location & Mood \\
    \midrule
    Examples &
    \makecell{\texttt{drinking},\\ \texttt{phoning}} &
    \makecell{\texttt{market},\\ \texttt{residential}} &
    \makecell{\texttt{focused},\\ \texttt{relaxed}} \\
    $|\YL|$ & 20 & 5 & 2 \\
    $|\YN|$ & 20 & 5 & 2 \\
    \midrule
    $|\DL|$ & 320 & 80 & 32 \\
    $|\DU|$ & 9.2K & 920 & 968 \\
    \bottomrule
    \end{tabular}
    \hspace{5pt}
    \begin{tabular}[b]{l cccc}
    \toprule
    \textbf{b) Clevr-4} & Texture & Color & Shape & Count \\
    \midrule
    Examples &
    \makecell{\texttt{metal},\\ \texttt{rubber}} &
    \makecell{\texttt{red},\\ \texttt{blue}} &
    \makecell{\texttt{torus},\\ \texttt{cube}} &
    \makecell{\texttt{1}, \texttt{2}} \\
    $|\YL|$ & 5 & 5 & 5 & 5 \\
    $|\YN|$ & 5 & 5 & 5 & 5 \\
    \midrule
    $|\DL|$ & 80 & 80 & 80 & 80 \\
    $|\DU|$ & 8.3K & 8.3K & 8.3K & 8.3K \\
    \bottomrule
    \end{tabular}%
\end{table*}

\begin{table}[h!]
    \centering\small
    \caption{\textbf{Class names for each dataset.} Classes in \textbf{bold} represent the known classes for the respective datasets.}
    \label{tab:class-names-supp}
    \begin{tabular}{@{}l|p{0.8\linewidth}@{}}
        \toprule
        Dataset & Class Names \\ \midrule
        Stanford Action & \textbf{applauding, brushing teeth, climbing, cutting trees, drinking, fishing, fixing a car, holding an umbrella, looking through a microscope, phoning, playing violin, pushing a cart, riding a bike, rowing a boat, shooting an arrow, taking photos, throwing frisby, walking the dog, watching TV, writing on a board}, blowing bubbles, cleaning the floor, cooking, cutting vegetables, feeding a horse, fixing a bike, gardening, jumping, looking through a telescope, playing guitar, pouring liquid, reading, riding a horse, running, smoking, texting message, using a computer, washing dishes, waving hands, writing on a book \\ \midrule
        Stanford Location & \textbf{educational institution, natural environment, office or workplace, public event or gathering, residential area}, restaurant or dining area, sports facility, store or market, transportation hub, urban area or city street \\ \midrule
        Stanford Mood & \textbf{adventurous, joyful}, focused, relaxed \\ \midrule
        Clevr-4 Texture & \textbf{rubber, metal, checkered, emojis, wave}, brick, star, circles, zigzag, chessboard \\ \midrule
        Clevr-4 Color & \textbf{gray, red, blue, green, brown}, purple, cyan, yellow, pink, orange \\ \midrule
        Clevr-4 Shape & \textbf{cube, sphere, monkey, cone, torus}, star, teapot, diamond, gear, cylinder \\ \midrule
        Clevr-4 Count & \textbf{7, 10, 1, 3, 5}, 2, 4, 6, 8, 9 \\
        \bottomrule
    \end{tabular}
\end{table}

\begin{table}[h!]
    \caption{\textbf{Omni accuracy evaluation set sizes for each dataset.} To compute the Omni accuracy, we gather all images labeled across all contexts for the evaluation set. For instance, this table shows that only 8 images overlap between the known image sets of the Action, Location, and Mood contexts in the Stanford dataset.}
    \label{tab:eval-size-supp}
    \centering\small
    \begin{tabular}[b]{l ccc}
    \toprule
    & Known & Novel & Overall \\
    \midrule
    Stanford & 8 & 17 & 128 \\
    Clevr-4  & 583 & 496 & 8,424 \\
    \bottomrule
    \end{tabular}
\end{table}

\subsection{Training of GCD and \sname}
\label{supp:training-details}

We begin each experiment using the exact training recipe from GCD~\cite{vaze2022generalized}. However, we find that the default hyperparameters lead to ineffective and unstable training due to the reduced number of labeled examples, the overall dataset size (Stanford \context{location}, \context{mood}), and out-of-distribution settings (Clevr-4). To address this, we perform
hyperparameter tuning directly on the unlabeled images in the training set for each dataset, based on the training loss curves and clustering quality based on the silhouette score. The silhouette score evaluates the quality of clustering by measuring how similar data points are within the same cluster compared to points in other clusters, which is an effective estimator of how well our model understands the current context and discovery of open categories. A separate validation set is also suboptimal for this task, as category discovery relies on the grouping of similar images, making dataset size critical. The hyperparameters used are detailed in \cref{tab:hyperparameter-supp}.

\begin{table}[h!]
    \caption{\textbf{Hyperparameters for training \sname. on Stanford and Clevr-4.} We start from GCD training recipe and perform unsupervised hyperparameter tuning based on training loss curves and clustering quality. CFJ = [RandomCrop, RandomHorizontalFlip, ColorJitter]. } 
    \label{tab:hyperparameter-supp}
    \centering\small
    \begin{minipage}{\linewidth}
    \centering\small
    \begin{tabular}[b]{l | ccc cccc}
    \toprule
        \multirow{2.5}{*}{Hyperparameter} &
        \multicolumn{3}{c}{Stanford} &
        \multicolumn{4}{c}{Clevr-4} \\
        \cmidrule(lr){2-4} \cmidrule(lr){5-8}
        &
        Action & Location & Mood &
        Texture & Color & Shape & Count \\
    \midrule
        batch\_size & 128 & 128 & 128 & 128 & 128 & 128 & 128 \\
        total\_epochs & 50 & 50 & 50 & 50 & 50 & 50 & 50 \\
        learning\_rate & 0.1 & 0.01 & 0.1 & 0.01 & 0.1 & 0.1 & 0.01 \\
        learning\_rate\_scheduler & Cosine & Cosine & Cosine & Cosine & Cosine & Cosine & Cosine \\
        min\_learning\_rate\_multiplier & 1e-3 & 1e-3 & 1e-3 & 1e-3 & 1e-3 & 1e-3 & 1e-3 \\
        optimizer & SGD & SGD & SGD & SGD & SGD & SGD & SGD\\
        momentum & 0.9 & 0.9 & 0.9 & 0.9 & 0.9 & 0.9 & 0.9 \\
        weight\_decay & 5e-5 & 5e-5 & 5e-5 & 5e-5 & 5e-5 & 5e-5 & 5e-5 \\
        context\_tokens\_length & 50 & 50 & 50 & 50 & 50 & 50 & 50 \\
        $\ell_\text{self-con}$: temperature & 1.0 & 1.0 & 1.0 & 1.0 & 1.0 & 1.0 & 1.0 \\
        $\ell_\text{self-con}$: n\_views & 2 & 2 & 2 & 2 & 2 & 2 & 2 \\
        $\ell_\text{self-con}$: augmentation & CFJ & CFJ & CFJ & CFJ & CFJ & CFJ & CFJ \\
        $\ell_\text{sup-con}$: $\lambda$ (loss weight) & 0.35 & 0.35 & 0.35 & 0.35 & 0.35 & 0.35 & 0.35 \\
        $\ell_\text{text-reg}$: $\lambda_\text{text-reg}$ (labeled, unlabeled) & (0.1, 0.01) & (1.0,1.0) & (1.0, 0.1) & (1.0,1.0) & (0.1, 0.1) & (0.1, 1.0) & (1.0, 1.0)\\
        SS-KMeans: n\_init & 10 & 10 & 10 & 10 & 10 & 10 & 10 \\
        SS-KMeans: tolerance & 1e-4 & 1e-4 & 1e-4 & 1e-4 & 1e-4 & 1e-4 & 1e-4 \\
        SS-KMeans: max\_iterations & 200 & 200 & 200 & 200 & 200 & 200 & 200 \\
    \bottomrule
    \end{tabular}%
    \end{minipage}
\end{table}

\textbf{Assumptions beyond GCD.}
We remark that \sname makes no additional distributional assumptions beyond GCD. Both methods assume the number of novel classes is known, though they are capable of estimating it. The only difference is access to a pool of class names, which can be generated by an LLM at minimal cost.

\textbf{Class names for GCD.}
Class names are assigned via Hungarian matching between predicted cluster IDs and true labels across all images, based on one-hot label distance for both GCD and Ours. This is only for visualization (\cref{fig:vis}), as true labels are unavailable in practice. Instead, our cluster names (\cref{supp:clustername}) are inferred by matching cluster centers of image embeddings with text embeddings of class names via CLIP using cosine similarity.

\subsection{LLM prompt for CLIP-ZS and \sname}
\label{supp:llm-prompts}

To adapt CLIP zero-shot methods for predicting novel classes, we generate potential novel class names using the publicly available ChatGPT. We provide the known class names, the number of novel classes required, and a specific prompt to ChatGPT, then use the generated responses as the discovered novel class names for zero-shot classification. \sname's text guidance algorithm and naming clustering algorithm for the unlabeled images follow a similar pipeline, with the key difference being that we request a significantly (up to 4 times) larger vocabulary from ChatGPT to construct our constrained vocabulary set. Our prompt is detailed below:

\begin{quote}
\emph{I have a dataset of images from the following classes: \textbf{[KNOWN\_CLASSES]}. What are the most possible classes that will also be included in this dataset? Give me \textbf{[NUMBER\_OF\_NOVEL\_CLASSES]} class names, only return class names separated by commas. Include quotation marks for each one.}
\end{quote}

\clearpage
\section{Additional saliency maps}
\label{supp:saliency}

\begin{figure*}[ht!]
\small\centering
\newcommand{\squareimage}[1]{\includegraphics[width=0.136\linewidth, height=0.136\linewidth]{#1}}
\newcommand{\colspace}{\hspace{5pt}}
\newcommand{\colspaceclose}{\hspace{3pt}}
\newcommand{\tablerow}[1]{
\squareimage{figs/saliency_supp/stanford_#1.png} &
\squareimage{figs/saliency_supp/stanford_#1_clip.png} &
\squareimage{figs/saliency_supp/stanford_#1_gcd_action.png} &
\squareimage{figs/saliency_supp/stanford_#1_cdc_action.png} &
\squareimage{figs/saliency_supp/stanford_#1_gcd_location.png} & 
\squareimage{figs/saliency_supp/stanford_#1_cdc_location.png} &
\squareimage{figs/saliency_supp/stanford_#1_gcd_mood.png} & 
\squareimage{figs/saliency_supp/stanford_#1_cdc_mood.png}
}
\resizebox{1.0\linewidth}{!}{

\begin{tabular}
{@{}
c@{\colspace}c@{\colspace}
c@{\colspaceclose}c@{\colspace}
c@{\colspaceclose}c@{\colspace}
c@{\colspaceclose}
c@{}}

\multicolumn{2}{l}{\textbf{a) Success cases}} &
\multicolumn{2}{c}{Action} &
\multicolumn{2}{c}{Location} &
\multicolumn{2}{c}{Mood} \\
\cmidrule(lr){3-4}\cmidrule(lr){5-6}\cmidrule(lr){7-8}
Image & CLIP &
GCD & \sname (ours) &
GCD & \sname (ours) &
GCD & \sname (ours) \\ [2pt]

\tablerow{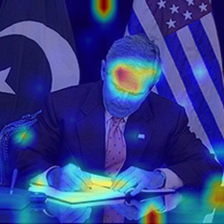}\\[-1pt]
&  & %
\textcolor{darkred}{reading} & \textcolor{darkgreen}{writing on a book} & 
\textcolor{darkred}{transportation hub} & \textcolor{darkgreen}{workplace} & 
\textcolor{darkgreen}{focused} & \textcolor{darkgreen}{focused} \\[2pt]
\tablerow{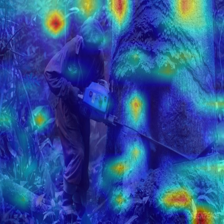}\\[-1pt]
& & %
\textcolor{darkgreen}{cutting trees} & \textcolor{darkgreen}{cutting trees} & 
\textcolor{darkgreen}{natural} & \textcolor{darkgreen}{natural} & 
\textcolor{darkgreen}{focused} & \textcolor{darkgreen}{focused} \\[2pt]
\tablerow{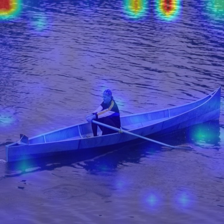}\\[-1pt]
& & %
\textcolor{darkgreen}{rowing a boat} & \textcolor{darkgreen}{rowing a boat} & 
\textcolor{darkgreen}{natural} & \textcolor{darkgreen}{natural} & 
\textcolor{darkred}{focused} & \textcolor{darkgreen}{adventurous} \\[8pt]

\multicolumn{2}{l}{\textbf{b) Failure cases}} &
\multicolumn{2}{c}{Action} &
\multicolumn{2}{c}{Location} &
\multicolumn{2}{c}{Mood} \\
\cmidrule(lr){3-4}\cmidrule(lr){5-6}\cmidrule(lr){7-8}
Image & CLIP &
GCD & \sname (ours) &
GCD & \sname (ours) &
GCD & \sname (ours) \\ [2pt]

\tablerow{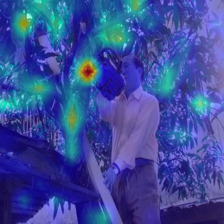}\\[-1pt]
& & %
\textcolor{darkgreen}{cutting trees} & \textcolor{darkred}{cleaning the floor} & 
\textcolor{darkgreen}{natural} & \textcolor{darkgreen}{natural} & 
\textcolor{darkgreen}{focused} & \textcolor{darkgreen}{focused} \\[2pt]
\tablerow{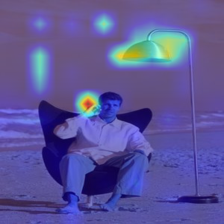}\\[-1pt]
& & %
\textcolor{darkgreen}{phoning} & \textcolor{darkgreen}{phoning} & 
\textcolor{darkred}{transportation} & \textcolor{darkred}{urban} (\textcolor{darkgreen}{natural}) & 
\textcolor{darkred}{joyful} & \textcolor{darkgreen}{relaxed} \\[2pt]
\tablerow{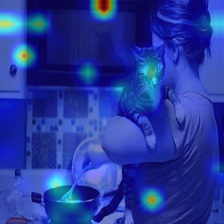}\\[-1pt]
& & %
\textcolor{darkred}{washing dishes} & \textcolor{darkgreen}{cooking} & 
\textcolor{darkgreen}{dining area} & \textcolor{darkgreen}{dining area} & 
\textcolor{darkred}{joyful} & \textcolor{darkred}{relaxed}  (\textcolor{darkgreen}{focused})\\ [2pt]

\end{tabular}
}
\caption{
\textbf{Additional saliency maps on the Stanford dataset} demonstrate that \sname makes reasonable predictions, focusing on the relevant regions for different contexts. We select three samples correctly predicted by \sname across all contexts and three that fail. In the failure cases, \sname 1) ignores the \textit{trees} with indirect interaction, mistaking the red saw for a \textit{cleaning} tool; 2) focuses on a lamp and a phone in a \textit{natural} beach scene, mistaking it for \textit{urban}; and 3) focuses on the \textit{relaxed} cat held by a \textit{focused} person closer to the camera.
}
\label{fig:vis-stanford-supp}
\end{figure*}

\clearpage

\begin{figure*}[t]
\small\centering
\newcommand{\squareimage}[1]{\includegraphics[width=0.105\linewidth, height=0.105\linewidth]{#1}}
\newcommand{\colspace}{\hspace{4pt}}
\newcommand{\colspaceclose}{\hspace{2pt}}
\newcommand{\tablerow}[1]{
\squareimage{figs/saliency_supp/clevr4_#1.png} & 
\squareimage{figs/saliency_supp/clevr4_#1_clip.png} & 
\squareimage{figs/saliency_supp/clevr4_#1_gcd_texture.png} &
\squareimage{figs/saliency_supp/clevr4_#1_cdc_texture.png} &
\squareimage{figs/saliency_supp/clevr4_#1_gcd_color.png} &
\squareimage{figs/saliency_supp/clevr4_#1_cdc_color.png} &
\squareimage{figs/saliency_supp/clevr4_#1_gcd_shape.png} &
\squareimage{figs/saliency_supp/clevr4_#1_cdc_shape.png} &
\squareimage{figs/saliency_supp/clevr4_#1_gcd_count.png} &
\squareimage{figs/saliency_supp/clevr4_#1_cdc_count.png} 
}
\resizebox{1.0\linewidth}{!}{
\begin{tabular}
{@{}
c@{\colspace}c@{\colspace}
c@{\colspaceclose}c@{\colspace}
c@{\colspaceclose}c@{\colspace}
c@{\colspaceclose}c@{\colspace}
c@{\colspaceclose}
c@{}}

\multicolumn{2}{l}{\textbf{a) Success cases}}&
\multicolumn{2}{c}{Texture} &
\multicolumn{2}{c}{Color} &
\multicolumn{2}{c}{Shape} &
\multicolumn{2}{c}{Count} \\
\cmidrule(lr){3-4}\cmidrule(lr){5-6}\cmidrule(lr){7-8}\cmidrule(lr){9-10}
Image & CLIP &
GCD & \sname (ours) &
GCD & \sname (ours) &
GCD & \sname (ours) &
GCD & \sname (ours) \\  [2pt]

\tablerow{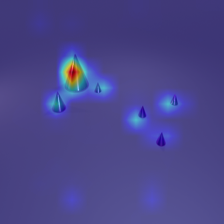}\\[-1pt]
& & %
\textcolor{darkred}{chessboard} & \textcolor{darkgreen}{metal} & 
\textcolor{darkgreen}{purple} & \textcolor{darkgreen}{purple} & 
\textcolor{darkgreen}{cone} & \textcolor{darkgreen}{cone} &
\textcolor{darkred}{5} & \textcolor{darkgreen}{6} \\ [2pt]
\tablerow{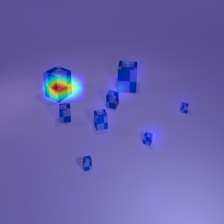}\\[-1pt]
& & %
\textcolor{darkred}{checkered} & \textcolor{darkgreen}{chessboard} & 
\textcolor{darkgreen}{orange} & \textcolor{darkgreen}{orange} & 
\textcolor{darkgreen}{cube} & \textcolor{darkgreen}{cube} &
\textcolor{darkred}{9} & \textcolor{darkgreen}{8} \\ [2pt]
\tablerow{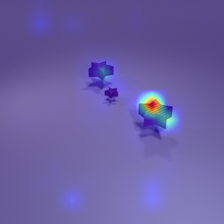}\\[-1pt]
& & %
\textcolor{darkred}{chessboard} & \textcolor{darkgreen}{wave} & 
\textcolor{darkgreen}{purple} & \textcolor{darkgreen}{purple} & 
\textcolor{darkgreen}{star} & \textcolor{darkgreen}{star} &
\textcolor{darkred}{2} & \textcolor{darkgreen}{3} \\ [8pt]

\multicolumn{2}{l}{\textbf{b) Failure cases}}&
\multicolumn{2}{c}{Texture} &
\multicolumn{2}{c}{Color} &
\multicolumn{2}{c}{Shape} &
\multicolumn{2}{c}{Count} \\
\cmidrule(lr){3-4}\cmidrule(lr){5-6}\cmidrule(lr){7-8}\cmidrule(lr){9-10}
Image & CLIP &
GCD & \sname (ours) &
GCD & \sname (ours) &
GCD & \sname (ours) &
GCD & \sname (ours) \\  [2pt]

\tablerow{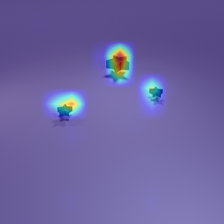}\\[-1pt]
& & %
\textcolor{darkred}{star} &  \textcolor{darkred}{star} (\textcolor{darkgreen}{brick}) & %
\textcolor{darkgreen}{star} & \textcolor{darkgreen}{star} & 
\textcolor{darkgreen}{brown} & \textcolor{darkgreen}{brown} &
\textcolor{darkred}{2} & \textcolor{darkgreen}{3} \\ [2pt]
\tablerow{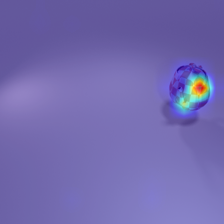}\\[-1pt]
& & %
\textcolor{darkgreen}{checkered} & \textcolor{darkgreen}{checkered} & 
\textcolor{darkgreen}{blue} & \textcolor{darkgreen}{blue} & 
\textcolor{darkred}{sphere} & \textcolor{darkred}{sphere} (\textcolor{darkgreen}{teapot})&
\textcolor{darkgreen}{1} & \textcolor{darkgreen}{1} \\ [2pt]
\tablerow{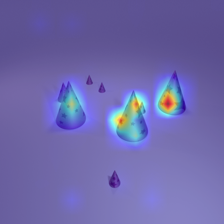}\\[-1pt]
& & %
\textcolor{darkred}{chessboard} & \textcolor{darkred}{metal} (\textcolor{darkgreen}{star})& 
\textcolor{darkgreen}{purple} & \textcolor{darkgreen}{purple} & 
\textcolor{darkgreen}{cone} & \textcolor{darkgreen}{cone} &
\textcolor{darkred}{5} & \textcolor{darkred}{6} (\textcolor{darkgreen}{8})\\ [2pt]

\end{tabular}
}
\caption{
\textbf{Additional saliency maps on the Clevr-4 dataset},
showing that \sname makes sensible predictions by focusing on relevant regions across various contexts, using the same setup. In the failure cases, \sname 1) struggles to identify black \textit{brick} patterns on a dark \textit{brown} object, mistaking the \textit{star} shape for a \textit{star} texture; 2) fails to recognize a \textit{teapot} at a challenging angle, mistaking it for a \textit{sphere}; and 3) has difficulty with smaller objects, leading to undercounting. Best viewed zoomed in.
}
\label{fig:vis-clevr-supp}
\end{figure*}

We present additional saliency maps for both the success cases and failure cases on the Stanford datasets and Clevr-4 datasets in \cref{fig:vis-stanford-supp} and \cref{fig:vis-clevr-supp}, respectively. \sname effectively switches between contexts, appropriately focusing on different aspects (regions) of the same image based on the context. Even in failure cases, the errors are easily interpretable and often arise from inherent ambiguities within the image, such as focusing on \textit{manufactured objects} like a lamp and a phone, leading to mispredicting \textit{nature} as \textit{urban}, as illustrated in the second failure example in the Stanford results.

\clearpage
\section{Full list of cluster names}
\label{supp:clustername}

We present the class names associated with known and novel visual clusters for Stanford \context{action}, \context{location}, \context{mood}, and Clevr-4 \context{texture}, \context{color}, \context{shape}, \context{count}, in
\cref{tab:clustername-stanford-action,tab:clustername-stanford-location,tab:clustername-stanford-mood,tab:clustername-clevr4-texture,tab:clustername-clevr4-color,tab:clustername-clevr4-shape,tab:clustername-clevr4-count}.
\sname identifies novel clusters accurately for contexts familiar to CLIP (Stanford \context{action}, \context{location} and Clevr-4 \context{color}, \context{shape}), but less accurate for unfamiliar contexts (Stanford \context{mood} and Clevr-4 \context{texture}, \context{count}) For familiar contexts, the predicted names align with synonyms, such as \keyword{preparing a meal} for \keyword{cooking} in Stanford \context{action} or \keyword{turquoise} for \keyword{cyan} in Clevr-4 \context{color}. Failure cases are also reasonable, such as predicting \textit{waving hands} as \textit{clapping}. In contrast, for unfamiliar contexts, the names are often unrelated, such as \keyword{admiring} for \keyword{relaxed} in Stanford \context{mood}.

\begin{table}[h]
    \centering\small
    \begin{minipage}[t]{0.48\textwidth} %
        \centering\small
        \caption{\textbf{Class names associated with every visual cluster from Stanford Action.} \sname's predictions largely align with the ground-truth labels provided by humans, often differing only in synonymous terms, with a few exceptions, such as the \textit{texting message} cluster being predicted as \textit{shaking hands}. Known classes are marked in \textbf{bold}. }
        \label{tab:clustername-stanford-action}
        \begin{tabular}{l l}
            \toprule
            GT Label & Prediction \\
            \midrule
            \textbf{applauding} & \textbf{applauding} \\
            \textbf{brushing teeth} & \textbf{brushing teeth} \\
            \textbf{climbing} & \textbf{rock climbing} \\
            \textbf{cutting trees} & \textbf{cutting trees} \\
            \textbf{drinking} & \textbf{drinking} \\
            \textbf{fishing} & \textbf{catching a fish} \\
            \textbf{fixing a car} & \textbf{fixing a car} \\
            \textbf{holding an umbrella} & \textbf{holding an umbrella} \\
            \textbf{looking through a microscope} & \textbf{looking in a microscope} \\
            \textbf{phoning} & \textbf{talking on a phone} \\
            \textbf{playing violin} & \textbf{playing violin} \\
            \textbf{pushing a cart} & \textbf{pushing a cart} \\
            \textbf{riding a bike} & \textbf{riding a bike} \\
            \textbf{rowing a boat} & \textbf{rowing a boat} \\
            \textbf{shooting an arrow} & \textbf{practicing archery} \\
            \textbf{taking photos} & \textbf{taking photos} \\
            \textbf{throwing frisby} & \textbf{fishing} \\
            \textbf{walking the dog} & \textbf{walking the dog} \\
            \textbf{watching TV} & \textbf{watching TV} \\
            \textbf{writing on a board} & \textbf{writing on a board} \\
            blowing bubbles & blowing bubbles \\
            cleaning the floor & mopping the floor \\
            cooking & preparing a meal \\
            cutting vegetables & climbing \\
            feeding a horse & petting a horse \\
            fixing a bike & fixing a bike \\
            gardening & weeding a garden \\
            jumping & dancing \\
            looking through a telescope & looking through a microscope \\
            playing guitar & strumming a guitar \\
            pouring liquid & carrying a box \\
            reading & reading a book \\
            riding a horse & running \\
            running & jogging \\
            smoking & smoking \\
            texting message & shaking hands \\
            using a computer & texting \\
            washing dishes & washing dishes \\
            waving hands & clapping \\
            writing on a book & writing a letter \\
            \bottomrule
        \end{tabular}
    \end{minipage}%
    \hfill
    \begin{minipage}[t]{0.48\textwidth} %
        \vspace{4em}
        \centering\small
        \caption{\textbf{Class names associated with every visual cluster from Stanford Location.} \sname's predictions often surpass the ground-truth labels in precision, capturing finer semantic meanings with greater granularity. We verify the correctness of these finer predictions through manual visual inspection. For example, many \textit{educational institutions} in our dataset are specifically \textit{science labs}, and many \textit{sports facilities} are \textit{rock climbing walls}. Known classes are marked in \textbf{bold}.}
        \label{tab:clustername-stanford-location}
        \begin{tabular}{ll}
            \toprule
            GT Label & Prediction \\
            \midrule
            \textbf{educational institution} & \textbf{science lab} \\
            \textbf{natural environment} & \textbf{natural environment} \\
            \textbf{office or workplace} & \textbf{office or workplace} \\
            \textbf{public event or gathering} & \textbf{public event or gathering} \\
            \textbf{residential area} & \textbf{residential area} \\
            restaurant or dining area & commercial kitchen \\
            sports facility & rock climbing wall \\
            store or market & road or highway \\
            transportation hub & language school \\
            urban area or city street & suburban street \\
            \bottomrule
        \end{tabular}
        \vspace{6em} %
        \caption{\textbf{Class names associated with every visual cluster from Stanford Mood.} Known classes are marked in \textbf{bold}.}
        \label{tab:clustername-stanford-mood}
        \begin{tabular}{ll}
            \toprule
            GT Label & Prediction \\
            \midrule
            \textbf{adventurous}  & \textbf{adventurous}  \\
            \textbf{joyful}  & \textbf{exhilarated}  \\
            focused  & explorative  \\
            relaxed  & admiring  \\
            \bottomrule
        \end{tabular}
    \end{minipage}
\end{table}

\begin{table}[ht]
    \centering\small
    \begin{minipage}[t]{0.48\textwidth} %
        \vspace{4em}
        \centering\small
        \caption{\textbf{Class names associated with every visual cluster from Clevr-4 Texture.} Known classes are marked in \textbf{bold}.}
        \label{tab:clustername-clevr4-texture}
        \begin{tabular}{ll}
            \toprule
            GT Label & Prediction \\
            \midrule
            \textbf{checkered} & \textbf{checkered} \\
            \textbf{emojis} & \textbf{emojis} \\
            \textbf{metal} & \textbf{metal} \\
            \textbf{rubber} & \textbf{rubber} \\
            \textbf{wave} & \textbf{wave} \\
            brick & abstract wave \\
            chessboard & chrome \\
            circles & checkerboard \\
            star & pixelated \\
            zigzag & wavy lines \\
            \bottomrule
        \end{tabular}
        \vspace{6em} %
        \caption{\textbf{Class names associated with every visual cluster from Clevr-4 Color.} Known classes are marked in \textbf{bold}.}
        \label{tab:clustername-clevr4-color}
        \begin{tabular}{ll}
            \toprule
            GT Label & Prediction \\
            \midrule
            \textbf{blue} & \textbf{indigo blue} \\
            \textbf{brown} & \textbf{warm brown} \\
            \textbf{gray} & \textbf{gray} \\
            \textbf{green} & \textbf{kelly green} \\
            \textbf{red} & \textbf{scarlet red} \\
            cyan & turquoise \\
            orange & orange \\
            pink & pink \\
            purple & lilac purple \\
            yellow & mustard yellow \\
            \bottomrule
        \end{tabular}
    \end{minipage}
    \hfill
    \begin{minipage}[t]{0.48\textwidth} %
        \vspace{4em}
        \centering\small
        \caption{\textbf{Class names associated with every visual cluster from Clevr-4 Shape.} Known classes are marked in \textbf{bold}.}
        \label{tab:clustername-clevr4-shape}
        \begin{tabular}{ll}
            \toprule
            GT Label & Prediction \\
            \midrule
            \textbf{cone} & \textbf{cone} \\
            \textbf{cube} & \textbf{cube} \\
            \textbf{monkey} & \textbf{monkey} \\
            \textbf{sphere} & \textbf{sphere} \\
            \textbf{torus} & \textbf{torus} \\
            star & star shape \\
            cylinder & cylinder \\
            diamond & diamond shape \\
            gear & gear \\
            teapot & teapot \\
            \bottomrule
        \end{tabular}
        \vspace{6em} %
        \caption{\textbf{Class names associated with every visual cluster from Clevr-4 Count.} Known classes are marked in \textbf{bold}.}
        \label{tab:clustername-clevr4-count}
        \begin{tabular}{ll}
            \toprule
            GT Label & Prediction \\
            \midrule
            \textbf{1} & \textbf{23} \\
            \textbf{3} & \textbf{3} \\
            \textbf{5} & \textbf{5} \\
            \textbf{7} & \textbf{7} \\
            \textbf{10} & \textbf{10} \\
            2 & 24 \\
            4 & 4 \\
            6 & 6 \\
            8 & 19 \\
            9 & 17 \\
            \bottomrule
        \end{tabular}
    \end{minipage}
\end{table}

\clearpage
\section{Additional results}
\label{supp:results}

\subsection{Results on standard benchmarks}
\label{supp:results-standard}

\sname also enhances GCD on standard single-context benchmarks by leveraging CLIP’s semantic knowledge and context-aware attention. \cref{tab:standard} shows full-shot results on CUB-200 and Stanford Cars using the CLIP ViT-B/16 backbone, demonstrating OAK’s superiority in novel class discovery. Moreover, \sname is compatible with state-of-the-art GCD methods and can be further improved by integrating them.

\begin{table}[h]
\centering\small
\caption{%
\textbf{Results on standard GCD benchmarks.}
}\label{tab:standard}
\begin{tabular}{lcaacaa}
\toprule
& \multicolumn{3}{c}{CUB-200} & \multicolumn{3}{c}{Stanford Cars} \\
\cmidrule(lr){2-4}\cmidrule(lr){5-7}
& Old & New & All & Old & New & All \\
\midrule
CLIP-ZS             & \textbf{69.4} & - & - & \textbf{81.4} & - & - \\
CLIP-ZS + LLM vocab & 46.4 & 44.0 & 44.8 & 54.6 & 47.4 & 49.7\\
CLIP-ZS + GT vocab  & 55.6 & 56.1 & 55.9 & 70.0 & 61.1 & 64.0\\
\midrule
SS-KMeans           & 46.2 & 46.6 & 46.5 & 51.1 & 43.5 & 46.0\\
GCD                 & 60.4 & 60.8 & 60.7 & 75.4 & 56.6 & 62.7 \\
\midrule
\sname (ours)       & 59.6 & \textbf{62.4} & \textbf{61.5} & 71.0 & \textbf{63.4} & \textbf{65.9} \\
\bottomrule
\end{tabular}
\end{table}

\subsection{Results on abstract textures}
\label{supp:results-abstract}

\begin{figure}[h]
\centering\small
\begin{minipage}{0.18\linewidth}
\includegraphics[width=\linewidth,height=\linewidth]{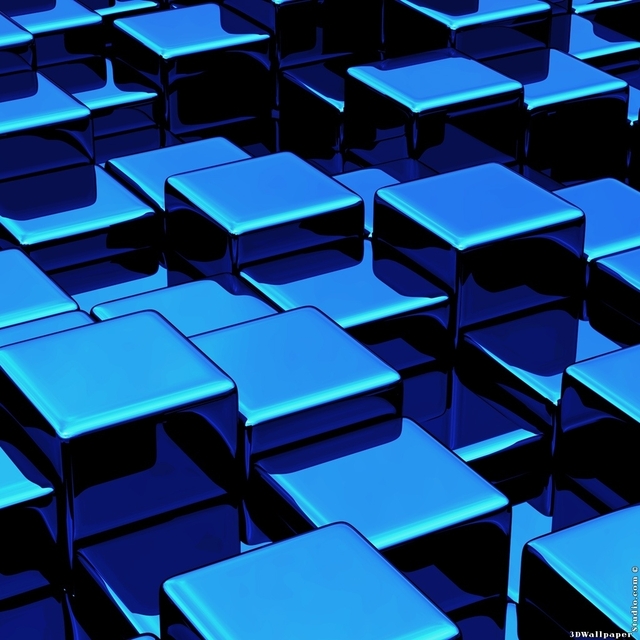}
\caption*{Known: \textit{bumpy}}
\end{minipage}%
~
\begin{minipage}{0.18\linewidth}
\includegraphics[width=\linewidth,height=\linewidth]{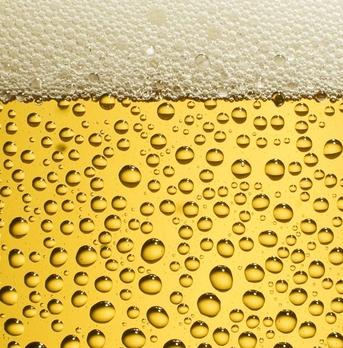}
\caption*{Novel: \textit{bubbly}}
\end{minipage}
\caption{%
\textbf{Example of known and novel classes in the DTD dataset.}
}\label{fig:dtd}
\end{figure}

We conduct experiments on the DTD~\citep{cimpoi2014describing} dataset, which contains images of abstract textures. Its 47 texture classes are split evenly into known and novel classes, using 20 labeled images per class. \cref{tab:dtd} shows that \sname outperforms the baselines, and \cref{fig:dtd} shows that \sname successfully discovers abstract classes such as \textit{bubbly}.

\begin{table}[h]
\centering\small
\caption{%
\textbf{Results on abstract textures.}
}\label{tab:dtd}
\begin{tabular}{lcaacaa}
\toprule
& Old & New & All \\
\midrule
CLIP-ZS             & 53.3 & - & -\\
CLIP-ZS + LLM vocab & 34.0 & 43.7 & 40.4\\
GCD                 & 55.4 & 61.7 & 59.6\\
\sname (ours)       & \textbf{56.7} & \textbf{65.0} & \textbf{62.1} \\
\bottomrule
\end{tabular}
\end{table}

\clearpage
\section{Additional analysis}
\label{supp:analysis}

\subsection{Ablation study}
\label{supp:ablation}

We study the effect of the components of \sname over \methodGcd on both the Stanford and Clevr-4 datasets in \cref{table:ablation-stanford} and \cref{table:ablation-clevr}. Consistent with prior observations, both context-aware attention and text-guided regularization contribute to the effectiveness of \sname, with their combination achieving the best results. This confirms the importance of integrating top-down text guidance and bottom-up image clustering to learn categorization rules. Specifically, context-aware attention~(\cref{eq:ctx-token}) enhances novel accuracy by adapting the visual encoder to new contexts, albeit sometimes at the cost of known accuracy, while top-down text guidance~(\cref{eq:loss-text}) improves known accuracy but may slightly reduce novel accuracy by constraining clusters with frozen text embeddings. When combined, our final \sname attains superior performance in both novel and overall accuracies, demonstrating the complementary nature of these two mechanisms. While CLIP-ZS provided limited benefit for synthetic images with abstract contexts in Clevr-4, leveraging text semantics significantly improved the overall accuracy of the baseline \methodGcd, particularly for higher-level contexts such as \context{texture} and \context{count}.

\begin{table*}[h!]
    \caption{%
    \textbf{Ablation study on the Stanford datasets.} Without context-aware attention and top-down text guidance, the first row largely represents the method of \methodGcd. With both critical components, the last row, \sname, outperforms the first row on the novel class performance, demonstrating the effectiveness of \sname.
    }\label{table:ablation-stanford}
    \centering\small
    \begin{minipage}{\linewidth}
    \centering\small
    \resizebox{1.0\linewidth}{!}{
    \begin{tabular}[b]{cc ccc>{\color{lightgray}}a ccca ccca}
    \toprule
        \multirow{2.5}{*}{\makecell{Context-aware\\attention}} &
        \multirow{2.5}{*}{\makecell{Top-down\\text guidance}} &
        \multicolumn{4}{c}{Known} &
        \multicolumn{4}{c}{Novel} &
        \multicolumn{4}{c}{Overall} \\
        \cmidrule(lr){3-6} \cmidrule(lr){7-10} \cmidrule(lr){11-14}
        & &
        Action & Location & Mood & Omni &
        Action & Location & Mood & Omni &
        Action & Location & Mood & Omni \\
    \midrule
    - & - &
    89.4 & 75.4 & 64.3 & 75.0 &
    67.8 & 80.8 & 40.6 & \phantom{0}0.0 &
    78.3 & 77.8 & 52.1 & 52.3 \\
    \checkmark & - &
    83.1 & 61.4 & 44.7 & \phantom{0}0.0 &
    73.3 & 75.0 & 51.8 & 23.5 &
    78.1 & 67.5 & 48.3 & 11.7 \\
    - & \checkmark &
    \textbf{95.7} & \textbf{87.6} & 66.9 & 50.0 &
    63.2 & 72.3 & 47.4 & \phantom{0}0.0 &
    79.0 & 80.0 & 56.8 & 43.8 \\
    \checkmark & \checkmark &
    88.9 & 83.9 & \textbf{68.8} & \phantom{0}0.0 &
    \textbf{85.1} & \textbf{88.4} & \textbf{87.4} & \textbf{47.1} &
    \textbf{86.9} & \textbf{85.9} & \textbf{78.4} & \textbf{70.3} \\
    \bottomrule
    \end{tabular}}
    \end{minipage}
\end{table*}

\begin{table*}[h!]
    \caption{%
    \textbf{Ablation study on Clevr-4} shows consistent results as those on the Stanford datasets, as shown in \cref{table:ablation-stanford}.
    }\label{table:ablation-clevr}
    \begin{minipage}{\linewidth}
    \centering\small
    \resizebox{1.0\linewidth}{!}{
    \begin{tabular}[b]{cc cccca cccca cccca}
    \toprule
        \multirow{2.5}{*}{\makecell{Context-aware\\attention}} &
        \multirow{2.5}{*}{\makecell{Text-guided\\regularization}} &
        \multicolumn{5}{c}{Known} &
        \multicolumn{5}{c}{Novel} &
        \multicolumn{5}{c}{Overall} \\
        \cmidrule(lr){3-7} \cmidrule(lr){8-12} \cmidrule(lr){13-17}
        & &
        Texture & Color & Shape & Count & Omni &
        Texture & Color & Shape & Count & Omni &
        Texture & Color & Shape & Count & Omni \\
    \midrule
    - & - &
    73.4 & 98.3 & 99.0 & 41.9 & 35.5 &
    43.6 & 94.9 & 99.2 & 42.3 & 15.7 &
    58.2 & 96.6 & 99.1 & 42.1 & 22.6 \\
    \checkmark & - & 
    35.0 & 99.5 & 98.9 & 39.2 & \phantom{0}8.2 &
    22.9 & 90.0 & 98.4 & 34.5 & \phantom{0}9.3 &
    28.8 & 94.7 & 98.7 & 36.9 & \phantom{0}7.6 \\
    - & \checkmark &
    74.8 & 98.1 & 99.4 & \textbf{52.3} & \textbf{53.9} &
    46.0 & 90.2 & 99.4 & 34.4 & 10.9 &
    60.1 & 94.1 & 99.4 & 43.3 & 27.9 \\
    \checkmark & \checkmark &
    \textbf{82.3} & \textbf{100.0} & \textbf{99.9} & 45.0 & 40.5 &
    \textbf{47.8} & \textbf{100.0} & \textbf{99.8} & \textbf{43.7} & \textbf{16.5} &
    \textbf{64.6} & \textbf{100.0} & \textbf{99.8} & \textbf{44.4} & \textbf{28.5}  \\
    \bottomrule
    \end{tabular}
    }
    \end{minipage}

\end{table*}

\newpage
\subsection{Multi-seed results}

We test the sensitivity of the 16 labeled images used for our final performance on the Stanford and CLEVR-4 datasets, applying five different random seeds for image selection in \cref{table:exp-stanford-multiseed} and \cref{table:exp-clevr-multiseed}, respectively. \sname consistently outperforms the baselines with statistical significance, achieving substantial margins beyond the standard deviations.

\begin{table*}[ht!]
    \centering\small
    \caption{%
    \textbf{Sensitivity analysis on the selection of 16 labeled images in the Stanford datasets.} We use five different random seeds for image selection, train GCD and \sname accordingly, and report the mean and standard deviation across the five runs. 
    }\label{table:exp-stanford-multiseed}
    \resizebox{.92\linewidth}{!}{
    \begin{tabular}[b]{l ccc>{\color{lightgray}}a ccca ccca}
    \toprule
        &
        \multicolumn{4}{c}{Known} &
        \multicolumn{4}{c}{Novel} &
        \multicolumn{4}{c}{Overall} \\
        \cmidrule(lr){2-5} \cmidrule(lr){6-9} \cmidrule(lr){10-13}
        Method &
        Action & Location & Mood & Omni &
        Action & Location & Mood & Omni &
        Action & Location & Mood & Omni \\
    \midrule
    \methodSSKmeans &
    63.0 & 62.8 & 25.9 & 12.5 &
    57.8 & 67.9 & \textbf{78.3} & 23.5 &
    60.3 & 65.1 & 52.9 & 22.6 \\
    &
    \stdv{4.2} & \stdv{7.1} & \stdv{0.6} & \stdv{0.0} &
    \stdv{3.5} & \stdv{4.7} & \stdv{0.2} & \stdv{4.2} &
    \stdv{1.5} & \stdv{3.8} & \stdv{0.4} & \stdv{4.1} \\
    \midrule
    \methodGcd &
    87.8 & 78.7 & 46.2 & 27.5 &
    62.1 & 78.4 & 46.1 & 17.6 &
    74.6 & 78.6 & 46.1 & 45.3 \\
    &
    \stdv{6.7} & \stdv{5.4} & \stdv{20.5} & \stdv{28.5} &
    \stdv{7.8} & \stdv{1.8} & \stdv{12.6} & \stdv{20.8} &
    \stdv{6.9} & \stdv{2.9} & \stdv{6.5} & \stdv{10.1} \\
    \midrule
    \sname (ours) &
    \textbf{89.8} & \textbf{84.2} & \textbf{59.6} & 5.0 & %
    \textbf{79.0} & \textbf{80.3} & 77.4 & \textbf{37.6} &
    \textbf{84.2} & \textbf{82.4} & \textbf{68.6} & \textbf{49.7} \\
    &
    \stdv{0.4} & \stdv{1.5} & \stdv{12.3} & \stdv{6.8} &
    \stdv{0.3} & \stdv{1.5} & \stdv{12.7} & \stdv{14.2} &
    \stdv{1.7} & \stdv{1.1} & \stdv{11.0} & \stdv{14.9} \\
    \bottomrule
    \end{tabular}}
\end{table*}

\begin{table*}[ht!]
    \caption{%
    \textbf{Sensitivity analysis on the selection of 16 labeled images in the Clevr-4 datasets}, following the same settings in \cref{table:exp-stanford-multiseed}.
    }\label{table:exp-clevr-multiseed}
    \begin{minipage}{\linewidth}
    \vspace{2pt}
    \centering\small
    \resizebox{1.0\linewidth}{!}{
    \begin{tabular}[b]{l cccca cccca cccca}
    \toprule
        &
        \multicolumn{5}{c}{Known} &
        \multicolumn{5}{c}{Novel} &
        \multicolumn{5}{c}{Overall} \\
        \cmidrule(lr){2-6} \cmidrule(lr){7-11} \cmidrule(lr){12-16}
        Method &
        Texture & Color & Shape & Count & Omni &
        Texture & Color & Shape & Count & Omni &
        Texture & Color & Shape & Count & Omni \\
    \midrule
    \methodSSKmeans &
    13.0 & 11.3 & 79.3 & 24.2 & 0.2 &
    13.6 & 12.1 & 78.7 & 15.3 & 0.2 &
    13.3 & 11.7 & 79.0 & 19.7 & 0.1 \\
    &
    \stdv{0.1} & \stdv{0.8} & \stdv{8.2} & \stdv{0.4} & \stdv{0.1} &
    \stdv{0.3} & \stdv{0.8} & \stdv{6.3} & \stdv{0.5} & \stdv{0.3} &
    \stdv{0.1} & \stdv{0.0} & \stdv{1.9} & \stdv{0.1} & \stdv{0.02} \\
    \midrule
    \methodGcd &
    47.4 & 76.3 & 98.0 & 43.0 & 32.0 &
    37.1 & 64.9 & 99.1 & 33.5 & 10.0 &
    42.1 & 70.5 & 98.5 & 38.2 & 18.4 \\
    &
    \stdv{27.4} & \stdv{25.2} & \stdv{3.3} & \stdv{8.2} & \stdv{11.4} &
    \stdv{9.2} & \stdv{32.9} & \stdv{1.1} & \stdv{6.0} & \stdv{4.8} &
    \stdv{17.9} & \stdv{27.4} & \stdv{1.7} & \stdv{6.5} & \stdv{6.7} \\
    \midrule
    \sname (ours) &
    \textbf{78.8} & \textbf{99.5} & \textbf{100.0} & \textbf{45.0} & \textbf{44.5} &
    \textbf{47.0} & \textbf{99.8} & \textbf{99.8} & \textbf{39.2} & \textbf{14.5} &
    \textbf{62.6} & \textbf{99.6} & \textbf{99.9} & \textbf{42.1} & \textbf{26.7} \\
    &
    \stdv{4.0} & \stdv{1.0} & \stdv{0.0} & \stdv{3.7} & \stdv{4.1} &
    \stdv{2.4} & \stdv{0.3} & \stdv{0.03} & \stdv{1.6} & \stdv{1.9} &
    \stdv{2.5} & \stdv{0.5} & \stdv{0.03} & \stdv{1.2} & \stdv{1.9} \\
    \bottomrule
    \end{tabular}
    }
    \end{minipage}
\end{table*}

\newpage
\subsection{Class names from large datasets}

Ad-hoc category discovery is an open-ended problem covering diverse custom contexts, making LLMs a natural choice since large datasets for these contexts are generally unavailable. Nevertheless, we compare our class names with those from the Kinetics~\citep{carreira2019short} dataset, which contains 700 \context{action} classes. \cref{tab:clustername-kinetics} shows that both produce similar novel class names when the candidate set is sufficiently large, such as \keyword{sweeping floor} vs. \keyword{cleaning the floor}.

\begin{table}[ht]
    \centering\small
    \caption{%
    \textbf{Comparison of predicted class names using candidate sets generated by GPT and those retrieved from Kinetics-700.}
    }\label{tab:clustername-kinetics}
    \begin{tabular}{l l l}
        \toprule
        GT Label & From ChatGPT-4o & From Kinetics-700 \\
        \midrule
        blowing bubbles & blowing bubbles & blowing bubble gum \\
        cleaning the floor & mopping the floor & sweeping floor \\
        cooking & preparing a meal & cooking egg \\
        cutting vegetables & climbing & cutting apple \\
        feeding a horse & petting a horse & petting horse \\
        fixing a bike & fixing a bike & fixing bicycle \\
        gardening & weeding a garden & digging \\
        jumping & dancing & high jump \\
        looking through a telescope & looking through a microscope & using a microscope \\
        playing guitar & strumming a guitar & playing guitar \\
        pouring liquid & carrying a box & pouring milk \\
        reading & reading a book & reading book \\
        riding a horse & running & riding or walking with horse \\
        running & jogging & jogging \\
        smoking & smoking & smoking \\
        texting message & shaking hands & texting \\
        using a computer & texting & assembling computer \\
        washing dishes & washing dishes & washing dishes \\
        waving hands & clapping & waving hand \\
        writing on a book & writing a letter & reading book \\
        \bottomrule
    \end{tabular}
\end{table}

\subsection{Additional analysis on Count}

\begin{figure}[h]
\centering\footnotesize
\includegraphics[width=\linewidth]{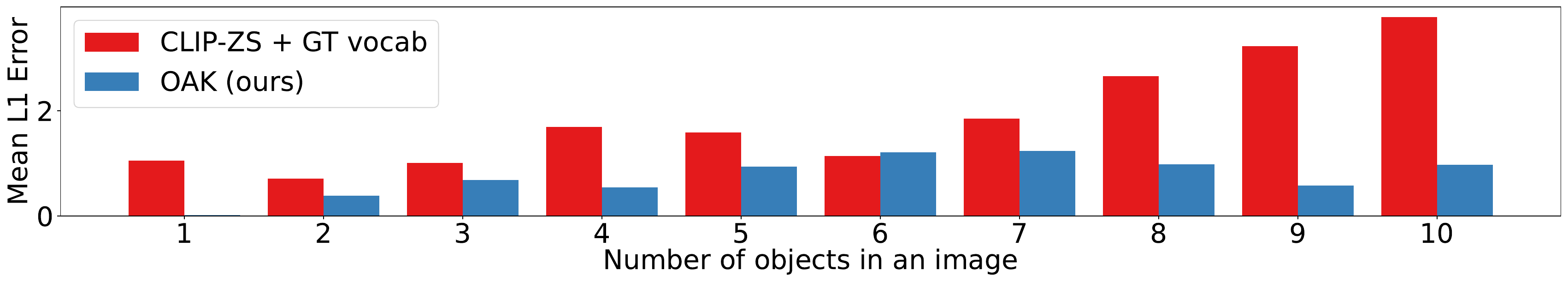}
\caption{%
\textbf{Mean error of OAK and CLIP versus the number of objects in an image.}
}\label{fig:count}
\end{figure}

We plot the mean error of \sname and CLIP against the number of objects in an image from the Clevr-4 dataset. For CLIP, we use the true names of novel classes, while \sname predicts them by matching cluster embeddings. \cref{fig:count} shows that CLIP struggles as object count increases, whereas \sname maintains stable performance. This highlights \sname’s ability to infer object counts through visual clustering, which is difficult to learn purely from semantics. Nonetheless, specialized object-counting models may still be needed for higher object counts ($>$10) beyond those in Clevr-4.

\newpage
\subsection{t-SNE visualizations}
\label{subsec:supp-tsne}

We present t-SNE plots of the feature spaces of CLIP and \sname on Stanford \context{action}, \context{location}, \context{mood}, and Clevr-4 \context{texture}, \context{color}, \context{shape}, \context{count} in the following figures. The plots show that \sname refines CLIP features into well-clustered representations aligned with each context. Notably, \sname performs well in contexts not familiar to CLIP, such as Stanford \context{location}. For out-of-distribution (OOD) images like Clevr-4 \context{shape} and \context{color}, \sname achieves near-perfect clustering. Even in cases that are both OOD and outside CLIP's primary focus, such as Clevr-4 \context{texture}, \sname forms reasonably coherent clusters, demonstrating its effectiveness. Note that the Clevr-4 \context{count} is a special case as the clusters learned by OAK shift gradually, reflecting smooth transitions between classes.

\begin{figure*}[ht!]
\centering
\includegraphics[width=\linewidth]{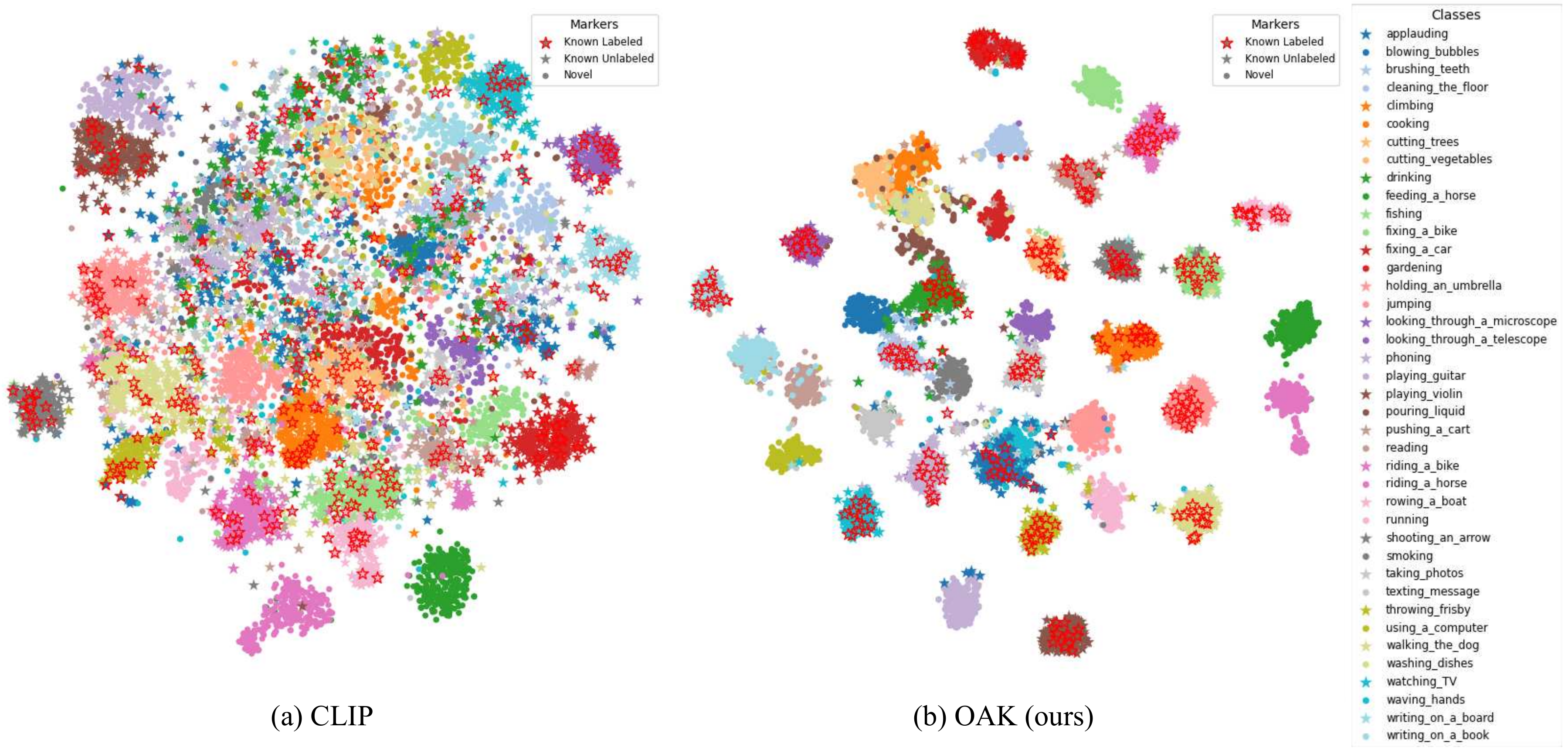}
\caption{%
\textbf{t-SNE plot of CLIP and \sname's feature space on Stanford \context{action}.}
}\label{fig:tsne-stanford-action-clip}
\end{figure*}

\begin{figure*}[ht!]
\centering
\includegraphics[width=\linewidth]{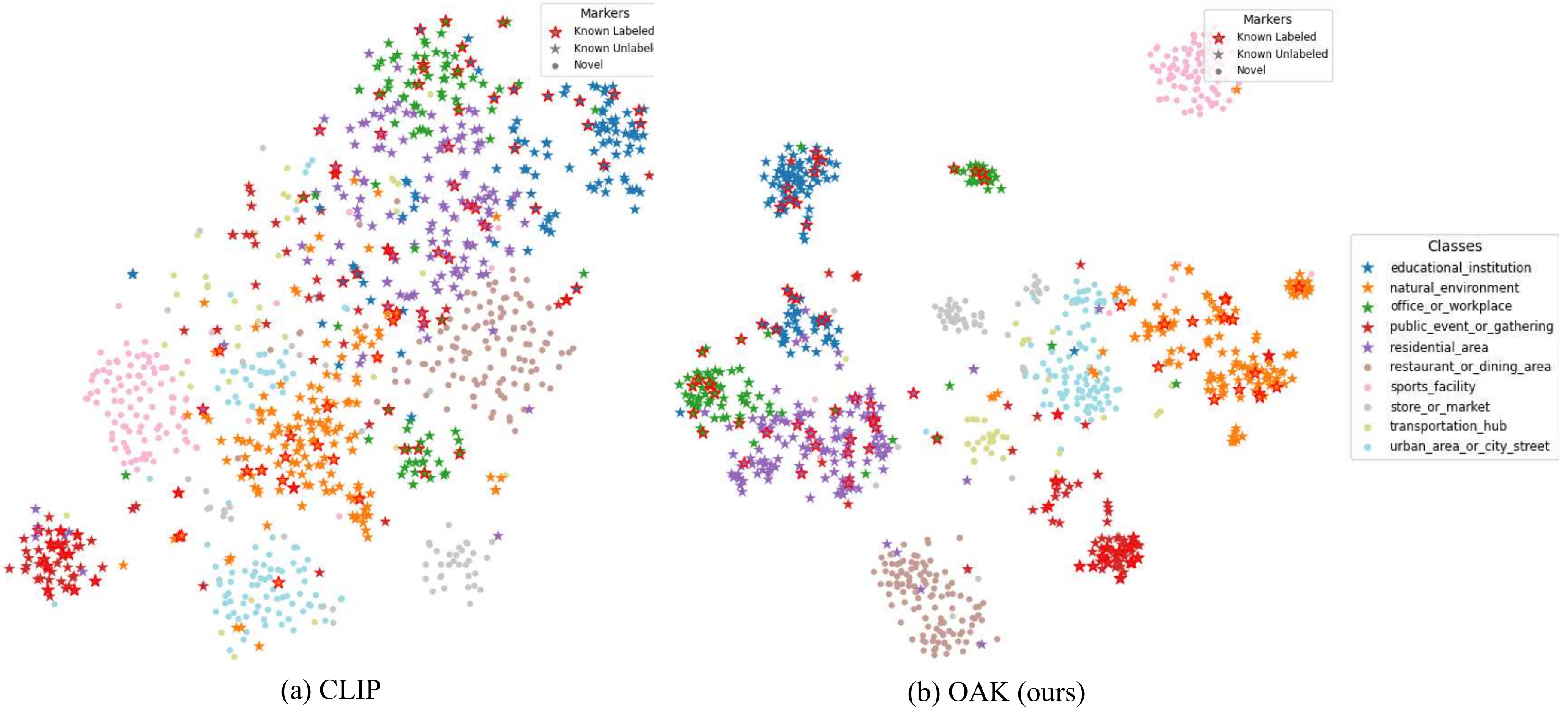}
\caption{%
\textbf{t-SNE plot of CLIP and \sname's feature space on Stanford \context{location}.}
}\label{fig:tsne-stanford-location-clip}
\end{figure*}

\begin{figure*}[ht!]
\centering
\includegraphics[width=\linewidth]{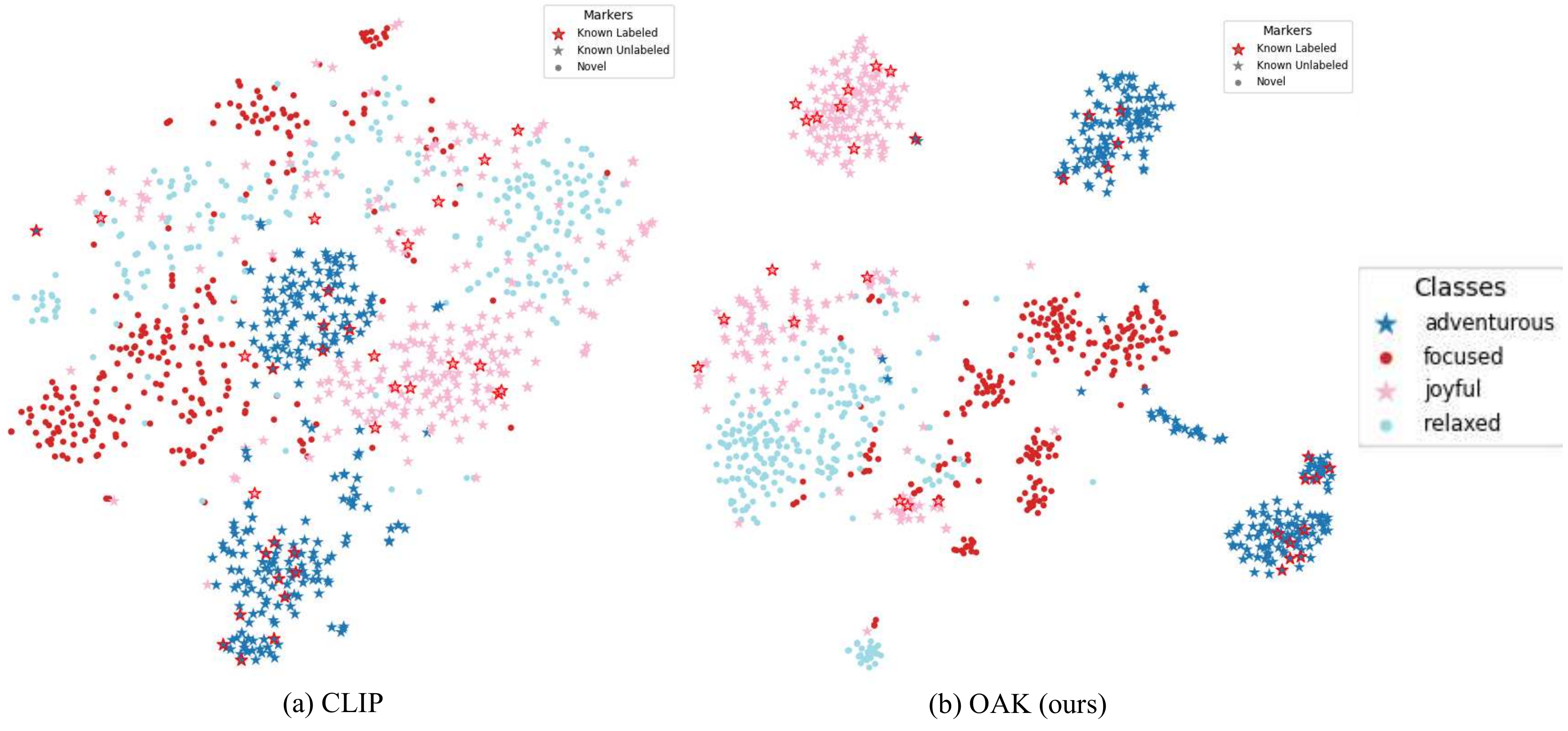}
\caption{%
\textbf{t-SNE plot of CLIP and \sname's feature space on Stanford \context{mood}.}
}\label{fig:tsne-stanford-mood-clip}
\end{figure*}

\begin{figure*}[ht!]
\centering
\includegraphics[width=\linewidth]{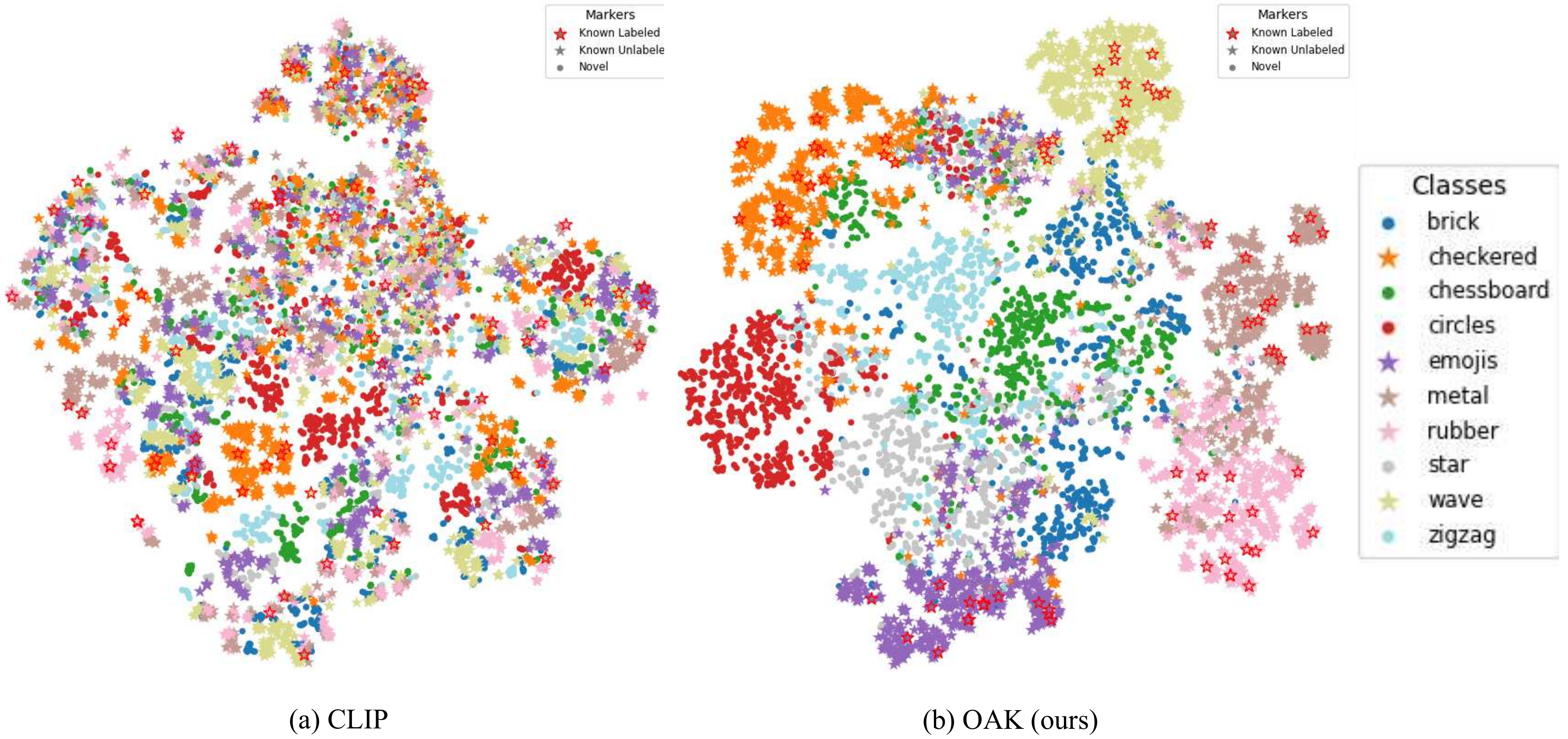}
\caption{%
\textbf{t-SNE plot of CLIP and \sname's feature space on CLEVR4 \context{texture}.}
}\label{fig:tsne-clevr4-texture-clip}
\end{figure*}

\begin{figure*}[ht!]
\centering
\includegraphics[width=\linewidth]{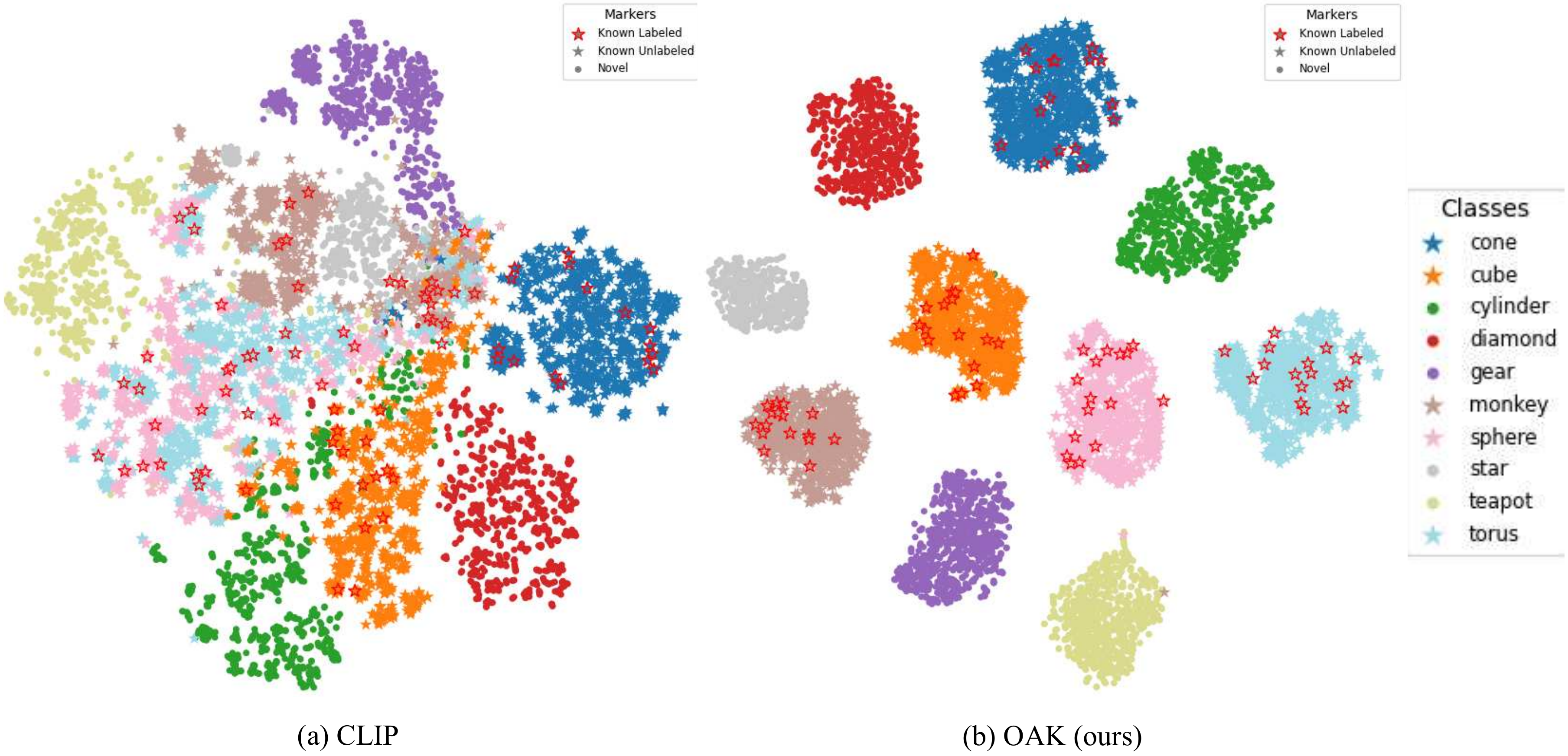}
\caption{%
\textbf{t-SNE plot of CLIP and \sname's feature space on CLEVR4 \context{shape}.}
}\label{fig:tsne-clevr4-shape-clip}
\end{figure*}

\begin{figure*}[ht!]
\centering
\includegraphics[width=\linewidth]{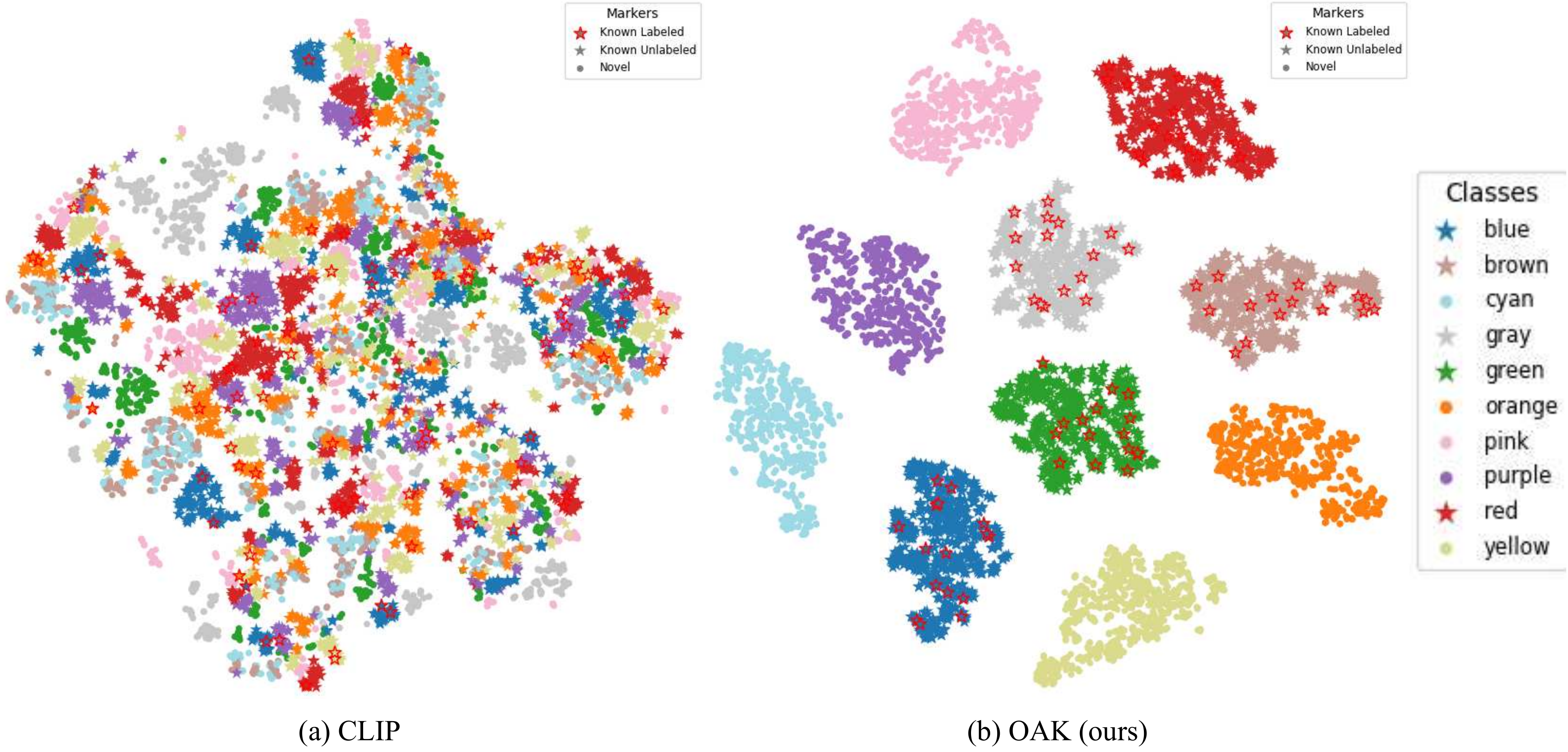}
\caption{%
\textbf{t-SNE plot of CLIP and \sname's feature space on CLEVR4 \context{color}.}
}\label{fig:tsne-clevr4-color-clip}
\end{figure*}

\begin{figure*}[ht!]
\centering
\includegraphics[width=\linewidth]{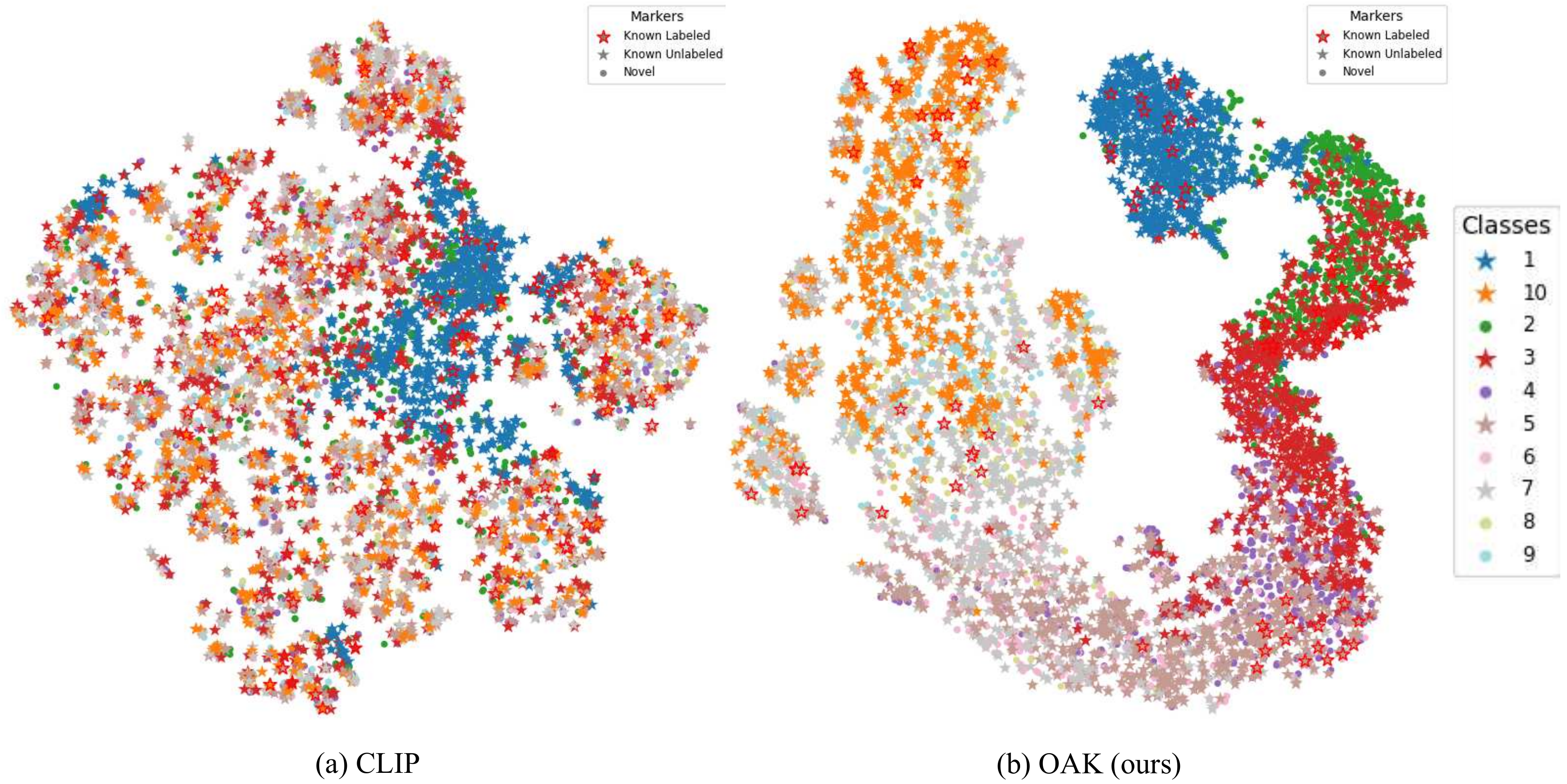}
\caption{%
\textbf{t-SNE plot of CLIP and \sname's feature space on CLEVR4 \context{count}.}
}\label{fig:tsne-clevr4-count-clip}
\end{figure*}

\end{document}